\documentclass[lettersize,journal]{IEEEtran}
\usepackage{amsmath,amsfonts}
\usepackage{algorithmic}
\usepackage{algorithm}
\usepackage{array}
\usepackage[caption=false,font=normalsize,labelfont=sf,textfont=sf]{subfig}
\usepackage{textcomp}
\usepackage{stfloats}
\usepackage{url}
\usepackage{verbatim}
\usepackage{graphicx}
\usepackage{cite}
\usepackage{hyperref}  
\usepackage{verbatim}  
\usepackage{adjustbox}
\usepackage{booktabs}
\usepackage{multirow}
\usepackage{amssymb}
\usepackage{bbding}
\usepackage{chronosys}
\usepackage{amsfonts}
\usepackage{makecell}
\begin{document}

\title{Delving into Multi-modal Multi-task Foundation Models for Road Scene Understanding: \\ From Learning Paradigm Perspectives}

\author{Sheng Luo, Wei Chen, Wanxin Tian, Rui Liu, Luanxuan Hou, Xiubao Zhang, Haifeng Shen, \\Ruiqi Wu, Shuyi Geng, Yi Zhou*, Ling Shao, Yi Yang, Bojun Gao, Qun Li and Guobin Wu
\thanks{
S. Luo, W. Chen, R. Wu, S. Geng and Y. Zhou are with the School of Computer Science and Engineering, Southeast University, Nanjing, China, and Key Laboratory of New Generation Artificial Intelligence Technology and Its Interdisciplinary Applications (Southeast University), Ministry of Education, China. Corresponding author: Yi Zhou (yizhou.szcn@gmail.com)\\
\indent W. Tian, R. Liu, L. Hou, X. Zhang, H. Shen, Y. Yang, B. Gao, Q. Li and G. Wu are with DiDi Chuxing, Beijing, China.\\
\indent L. Shao is with the UCAS-Terminus AI Lab, University of Chinese Academy of Sciences, Beijing, China}%
}

\maketitle

\begin{abstract}
Foundation models have indeed made a profound impact on various fields, emerging as pivotal components that significantly shape the capabilities of intelligent systems.  In the context of intelligent vehicles, leveraging the power of foundation models has proven to be transformative, offering notable advancements in visual understanding. Equipped with multi-modal and multi-task learning capabilities, multi-modal multi-task visual understanding foundation models (MM-VUFMs) effectively process and fuse data from diverse modalities and simultaneously handle various driving-related tasks with powerful adaptability, contributing to a more holistic understanding of the surrounding scene. In this survey, we present a systematic analysis of MM-VUFMs specifically designed for road scenes. Our objective is not only to provide a comprehensive overview of common practices, referring to task-specific models, unified multi-modal models, unified multi-task models, and foundation model prompting techniques, but also to highlight their advanced capabilities in diverse learning paradigms. These paradigms include open-world understanding, efficient transfer for road scenes, continual learning, interactive and generative capability. Moreover, we provide insights into key challenges and future trends, such as closed-loop driving systems, interpretability, low-resource conditions, embodied driving agents, and world models.
To facilitate researchers in staying abreast of the latest developments in MM-VUFMs for road scenes, we have established a continuously updated repository at \href{https://github.com/rolsheng/MM-VUFM4DS.git}{https://github.com/rolsheng/MM-VUFM4DS}. 

\end{abstract}


\section{Introduction}

\begin{figure*}[t]
    \centering
    \includegraphics[width=\linewidth]{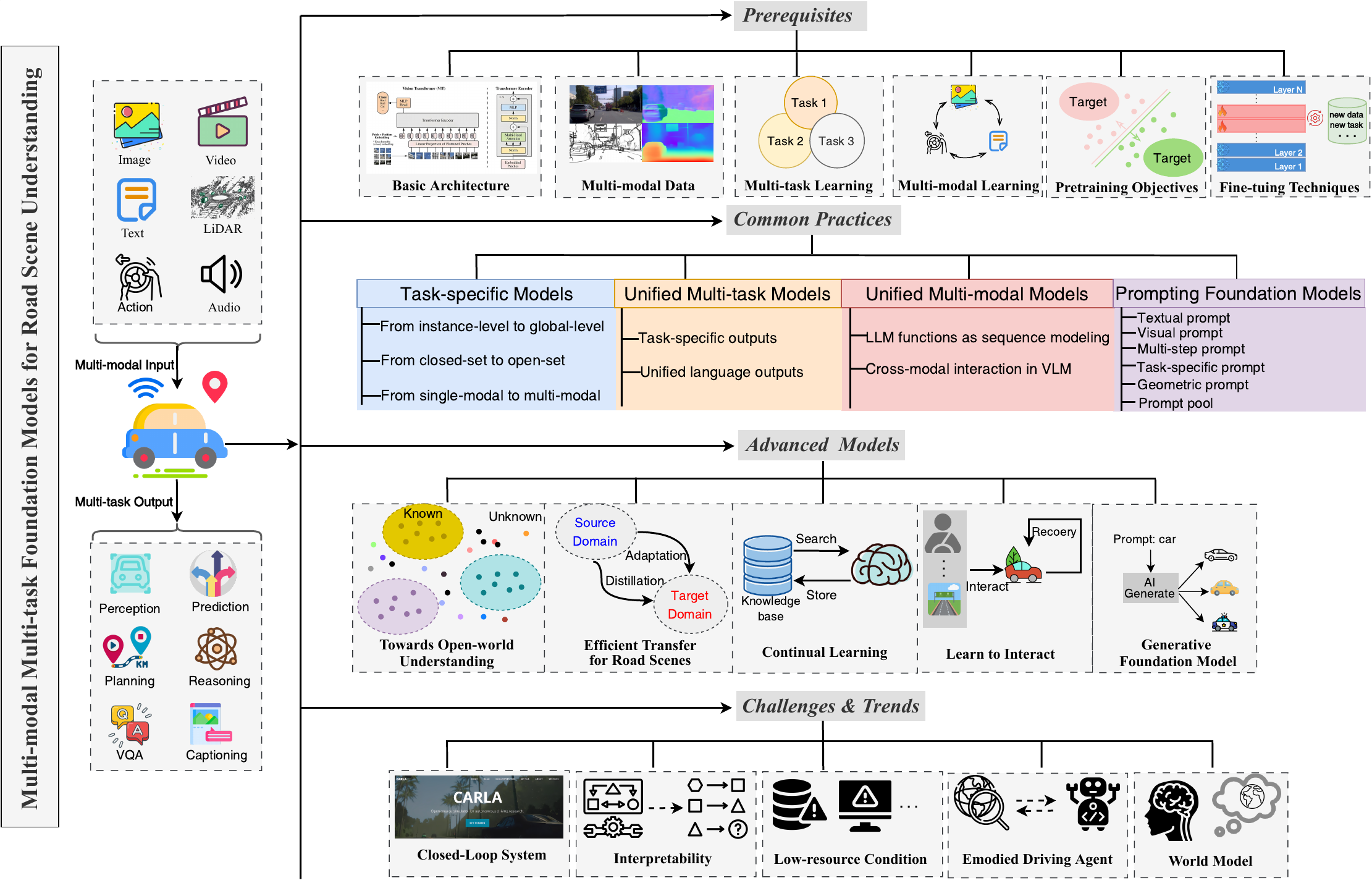}
    \caption{Overview of our survey at a glance. A multi-modal and multi-task foundation model for road scene understanding is defined as a framework that inputs multi-modal data and outputs multi-task results. In the section of prerequisites, we introduce some basic knowledge in advance before reading the main context. Then, we refer to up-to-date task-specific models, unified multi-task models, unified multi-task models for road scene understanding and prompting foundation models, respectively, in the section of common practice. The section of advanced models is to show strengths in diverse learning paradigms, such as open-world understanding, efficient transfer for road scene, continual learning, interactive and generative capabilities, respectively. Finally, we also list key challenges and promising future trends to address them.}
    \label{fig:overview}
\end{figure*}

 \IEEEPARstart{I}{ntelligent} vehicles have made great progress by achieving significant advancements in perceiving road scenes, employing various tasks such as object detection, trajectory prediction, and advanced generation methods. The integration of vision-centric (such as camera, Lidar) and vision-beyond (such as text, action) modalities further enhances the ability of autonomous vehicles to perceive the world from diverse dimensionalities. 
However, traditional driving-related models are typically designed for specific tasks and utilize common convolution neural networks (CNNs)\cite{resnet} or Transformer\cite{vit, detr, setr} architecture to extract feature maps from a single modality. This fashion often leads to incomplete observation due to complex and unpredictable conditions in real-world road scenes.
To address this problem, researchers have shifted their focus to unified multi-modal and multi-task models, as illustrated in Fig.~\ref{fig:overview}. These models possess multi-task learning capabilities, allowing them to concurrently perform multiple tasks. Moreover, with the integration of multi-modal capabilities (e.g., various visual sensors, text), these models contribute to creating more versatile systems.

The emergence of Foundation Models (FMs) is regarded as a milestone in achieving artificial general intelligence (AGI). Recently, large language models (LLMs) \cite{gpt-3,gpt-4,chatgpt,llama,llama2,instructgpt}, vision language models (VLMs)\cite{visualchatgpt,minigpt4,instructblip,videollama,llava_improved,llava}, and large vision models (LVMs) \cite{lvm_sequential_modeling,sam}, have already drawn significant attention. LLMs require training on a large amount of textual data and considerable computational resources, showing powerful emergent abilities with continuous growth of data scale. With the extensive world knowledge from the pretraining stage, LLMs have been viewed as knowledge bases to solve downstream tasks by prompt engineering. VLMs refer to a series of models designed to bridge the gap between visual information and natural language understanding (e.g., Vision Question Answer(VQA), Captioning), emphasizing on aligning vision modality into language latent space and thereby enhancing the capacity to understanding and reasoning about visual content. Moreover, LVMs aim to achieve powerful strengths in vision-centric tasks without relying on linguistic data. These FMs with exceptional generalization capabilities have achieved great success across various domains. 

\textbf{Motivation of our survey.} Among traditional algorithms for road scene understanding, each algorithm is individually designed to solve a specific task, and is usually trained on a single modality. This is beneficial for easy assembly and deployment. However, we argue that these algorithms are inefficient and impractical in real-world scenarios. 1) First, limited knowledge can be learned from a single task, hindering mutual benefit from learning universal knowledge across multiple tasks. 2) Second, the acquisition and processing of multi-modal data have become increasingly feasible due to the advancement of multi-modal learning and the flexibility of Transformer-based architecture, respectively. By integrating these multi-modal data into a unified multi-modal model, we can gain comprehensive understanding of road scene characteristics. 3) Third, the intricacies of road scenes demand systems that possess a profound understanding of the surrounding environment. Motivated by the great success of FMs, the realization of their pivotal role in the visual understanding of road scenes has taken center stage. The great generalization capabilities of MM-VUFMs make them an ideal solution to tackle this issue. Due to the emergence of numerous related research papers and the absence of a review in the current landscape, there is an urgent need for a comprehensive and systematic survey for MM-VUFMs.

\textbf{Comparisons with related surveys.} 
We investigated previous related surveys to clarify a difference between our survey with them\cite{visual_foundatio_models,knowledgedriven_AD,survey_applications,survey_datasets1,survey_e2e_dl1,survey_e2e_challenges,survey_llm,survey_mllm,survey_vlm,survey_foundation_model}. Some previous surveys \cite{survey_e2e_challenges,survey_mllm,survey_e2e_dl1,survey_llm,survey_vlm} cover related datasets, methods, and applications from a perspective of end-to-end autonomous driving (e.g., perception, prediction, and planning), causing a lack of detailed attention to visual understanding. The survey \cite{visual_foundatio_models} focuses on specific views such as data generation,self-supervised learning, and adaptation, but they have no consideration of multi-modal and multi-task capabilities of FMs. However, our survey not only reviews the latest works of MM-VUFMs from the perspective of multi-modal and multi-task capabilities but also has more emphasis on their advanced strength in diverse learning paradigms, aiming to provide readers with comprehensive awareness and in-depth insights toward this field.

\textbf{Contributions.} In the end, we summarize four contributions of this survey as follows:
\begin{itemize}
\item We provide a systematic analysis of up-to-date (until May, 2024) MM-VUFMs for road scenes, covering high-level motivation, common practices, advanced capabilities in diverse learning paradigms, emerging challenges and future trends.
\item We classify existing MM-VUFMs into task-specific models, unified multi-task, unified multi-modal models, and foundation model prompting techniques, respectively. We also review datasets for road scenes from multi-modal and multi-task perspectives, and typical evaluation metrics for measuring performance.
\item Advanced strengths of MM-VUFMs in diverse learning paradigms are highlighted, including open-world understanding, efficient transfer, low-resource condition, continual learning, interactive and generative capability.
\item  Emerging key challenges and promising trends are proposed to draw attention to achieve intelligent vehicles.
\end{itemize}

\section{Prerequisites and Roadmap}

\subsection{Prerequisites}

\subsubsection{Basic Architectures}  \par
Convolutional neural networks (CNNs)\cite{resnet} are classic architectures for computer vision tasks, opening a new era for computer vision and continuing to maintain their prominence and impact in the era of large-scale foundation models\cite{clip, internimage, vit+cnn_1}. Despite CNN's long-range dependencies through applying large kernels or recursive gated kernels, the performance is still limited in scaling up in-context learning.
Transformer\cite{transformer} is an attention-based sequence-to-sequence learning architecture originally introduced to extract long-range dependencies from natural language. Unlike previous models that relied on recurrent or convolutional layers, Transformer exclusively employs the self-attention\cite{transformer} mechanism to weigh different parts of the input sequence, enabling parallelization and improving efficiency in capturing long-range dependencies.


Vision-Language architecture\cite{clip,flamingo,blip2}. The success of Transformer in both language and vision domains has spurred the development of vision-language architecture. This architecture leverages Transformers' ability to handle serialized multi-modal data. To be more specific, they often employ mechanisms such as cross-attention, which functions by enabling queries from one modality(e.g., text) to interact with keys and values from another modality(e.g., image), facilitating a context-rich understanding that is essential for tasks requiring joint visual and textual comprehension such as image captioning, visual question answering (VQA).

\subsubsection{Multi-modal Data}
\begin{figure}[t]
    \centering
    \includegraphics[width=\linewidth]{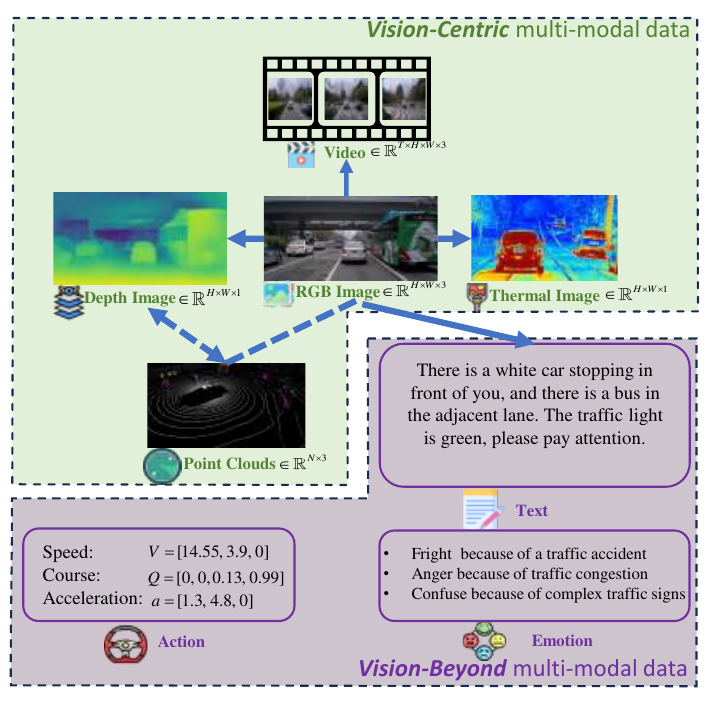}
    \caption{Common multi-modal data used in road scenes. We divide them into two groups, i.e. \textbf{vision-centric} multi-modal data and \textbf{vision-beyond} multi-modal data. Solid arrows denote the strong connections between two modalities and dashed arrows denote weak connections. Vision-centric multi-modal data refer to those collected from perception sensors, usually containing detailed visual features, while vision-beyond multi-modal data refer to those springing up recently which contain more semantic and comprehensive information describing the holistic scene.}
    \label{fig:multimodal-data}
\end{figure}

In road scenes, multi-modal data play a critical role in enhancing the perception, decision-making, and interaction capabilities of intelligent vehicles. These data come from various sources and types, each contributing unique information essential for the comprehensive functioning of road scene understanding \cite{multimodalfusion1}. Here we introduce several mainstream multi-modal data used in road scenes in Fig. \ref{fig:multimodal-data}. Based on their data modality, we categorize them into two groups: \textbf{vision-centric} multi-modal data which include the most popular used sensor data in the past decades, and \textbf{vision-beyond} multi-modal data that include other types of multi-modal data springing up in recent few years.
\par
  
\par 
Vision-centric multi-modal data include images and videos captured by various cameras such as RGB cameras, depth cameras, thermal cameras and lidar. RGB image $I$ can be represented as $I\in \mathbb{R}^{H\times W\times C}$ where $H$ and $W$ stand for the height and width of the image respectively. The symbol $C$ represents the number of channels: 3 for RGB images, and 1 for depth or thermal images. 
A sequence of images forms a video $V\in\mathbb{R}^{T\times H\times W\times 3}$ where $T$ stands for the number of frames. Depth images measure the distance between the object and the ego-car, and thermal images help detect moving objects with relatively high temperature. 
Point clouds from the lidar, featuring robustness to various weather conditions, can be represented as $P\in \mathbb{R}^{N\times 3}$ where $N$ denotes the number of points and 3 stands for spatial coordinates for each point.

Recently, there has been a growing trend to harness a broader spectrum of modalities which are called vision-beyond multi-modal data. In our survey, vision-beyond multi-modal data include text, emotion, and action. Text data provide an overview of the road scene using natural language descriptions. Emotion data are collected and annotated, reflecting the driver's emotions according to the real-time road scene. Action data are numerical data collected from the control system, demonstrating the moving status of the vehicle. The incorporation of these vision-beyond multi-modal data enables systems to gain more comprehensive scene understanding.

\subsubsection{Multi-modal learning}
The goal is to train models through the joint learning of multiple representations from diverse data modalities, including image, text, and video, mimicking the human ability to interact with the environment through various senses. Two primary methods for encoding multiple modalities are as follows:

\textbf{Modal-invariant encoder.} These methods map all modalities into a joint embedding space by a shared encoder. They tend to propose a simple and effective architecture to learn the ability to interact across modalities. The success of unified multi-modal learning methods such as Meta-Transformer \cite{meta-transformer} has inspired methods to learn joint embedding space by utilizing a unified multi-modal encoder. Meta-Transformer benefits from the versatility of Transformer architecture to learn modal-invariant representations. 

\textbf{Modal-specific encoder.} These methods use a modal-specific encoder for each modality. CLIP\cite{clip} individually utilizes two encoders for vision and language, achieving strong zero-shot learning toward downstream tasks by aligning visual-semantic space into language space via contrastive pretraining. ImageBind\cite{imagebind} involves more modalities such as video, audio, thermal data. By aligning modal-specific embedding into the image embedding, a joint embedding space for all modalities is obtained.
\subsubsection{Multi-task learning}
Multi-task learning (MTL) refers to a paradigm where a model is trained to perform multiple related tasks simultaneously. Instead of training separate models for individual task, MTL aims to leverage shared representations to improve overall performance on all tasks.

\textbf{Multi-task architectures.} Parameter sharing is a key mechanism to encourage the learning of shared information among different tasks. Depending on the extent of parameter sharing within model, multi-task architectures involve two categories: hard-parameter sharing\cite{kendall2018multi} and soft-parameter sharing\cite{mmoe}. Specifically, in hard-parameter sharing architecture, a single set of shared parameters or a shared feature extractor is used for all tasks to save computational resources. The shared parameters are responsible for extracting features that are considered task-agnostic. However, each task has its own task-specific parameters, which are responsible for generating task-specific outputs.
Unlike a strictly common set of shared parameters in hard-parameter sharing, the soft-parameter sharing architecture introduces a degree of flexibility by allowing task-specific parameters to be influenced by shared parameters. This architecture aims to strike a balance between leveraging shared knowledge and allowing flexibility for task-specific outputs.

\textbf{Multi-objective optimization.} During joint training, different tasks often correspond to different objectives. Multi-objective optimization aims to find a set of solutions that optimize multiple objectives simultaneously. One common approach involves transforming the multi-objective problem into a single-objective problem using a weighted sum of the objectives\cite{kendall2018multi,dynamic_weight}, which allows to express the relative importance or priority of each task during the optimization process. 
However, this fashion not only struggles to accurately capture the trade-offs between multiple conflicting objectives where the Pareto front is non-convex or has intricate shapes, but the performance is highly sensitive to the choice of these weights. Gradient-based multi-objective optimization is another valuable approach to address complex problems with conflicting objectives. For example, Pareto-based methods\cite{pareto,ma2020efficient,multitaskmultiobj} are often employed in gradient-based multi-objective optimization. These methods aim to find solutions along the Pareto front, which represents the set of optimal solutions that cannot be improved in one objective without degrading another, by iteratively updating the model parameters using gradients from multiple objectives.

\begin{figure*}[t]
    \centering
    \includegraphics[width=\linewidth]{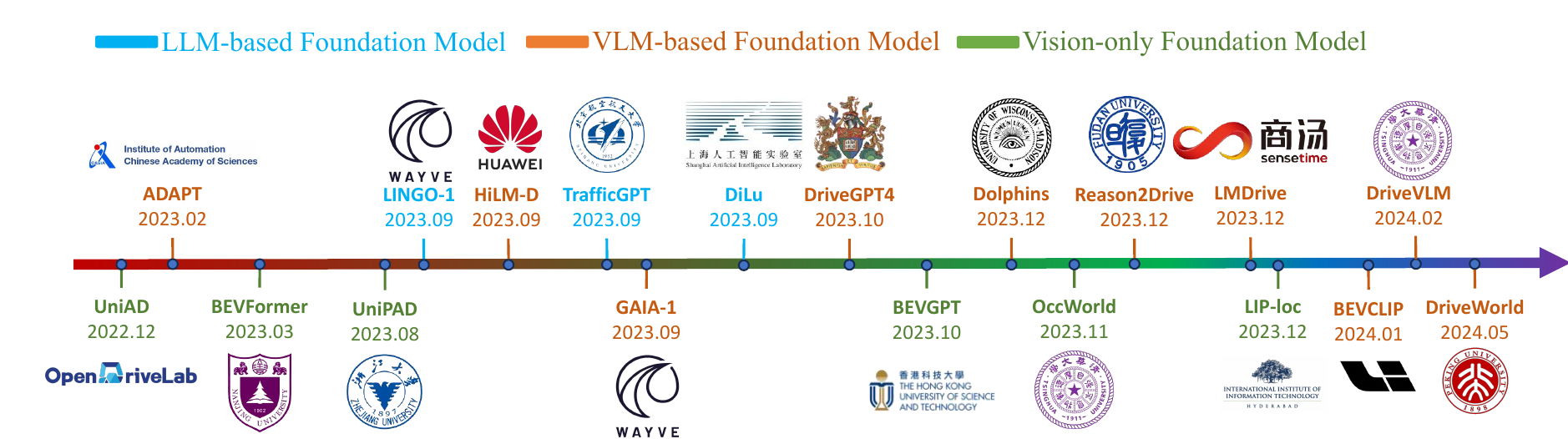}
    \caption{Roadmap of recent foundation models in driving scenarios. We divide these foundation models into LLFM, VLFM, and LVFM based on the data modality they use. LVFMs are vision-only large-vision foundation models that only take vision-centric data as input. Pretrained on large-scale datasets, these foundation models can act as robust feature representors and facilitate downstream tasks to a great extent. In contrast, LLFMs and VLFMs usually incorporate LLMs or VLMs respectively, leveraging their robust reasoning ability to perform various complicated tasks.}
    \label{fig:roadmap}
\end{figure*}
\subsubsection{Pretraining objectives}
Foundation models have propelled the fundamental cognitive abilities of intelligent systems. These models are pretrained on extensive datasets primarily through self-supervised or weakly-supervised learning methods. Through the stage of pretraining, the model can acquire the intrinsic structure and patterns from data. We summarized four types of pretraining methods as follows:

\textbf{Image-only contrastive pretraining.} To acquire a robust encoder, image-only contrastive pretraining builds on a specific pretext task where the model learns to compact positive sample pairs and widen negative sample pairs in a unified representation space. Existing methods \cite{moco,mocov2,simclr,simclrv2} usually construct pairs of positive and negative samples using various hand-craft augmentation of the same image, such as color jittering, crop and resize operations.
\begin{equation}
\label{image-only contrative loss}
  L_{image-only} = -\frac{1}{N}\sum_{i=1}^{N}\log\frac{\textit{exp}(x_{i}^{+}\cdot x_{0}/\tau)}{\sum_{j=1,j\neq i}^{N}\textit{exp}(x_{j}^{-} \cdot x_{0}/\tau)}.
\end{equation}
Specifically, they use a similar objective function like InfoNCE, as shown in Eq.~\ref{image-only contrative loss}, to maximize agreement between positive pairs and minimize agreement between negative pairs. Given a query sample $x_{0}$, $x^{+}$ and $x^{-}$ are individually positive and negative samples associated with $x_{0}$. $N$ is the total number of samples and $exp(\cdot)$ is similarity function.

\textbf{Image-Text contrastive pretraining.} With the popularity of multi-modal learning, researchers have also explored its potential for aligning image-text representation\cite{clip,align,k-lite,less,basic,beit}. The image-text contrastive loss is calculated as Eq.~\ref{image-text contrative loss}:
\begin{equation}
\label{image-text contrative loss}
  L_{image-text} = -\frac{1}{N}\sum_{i=1}^{N}\log\frac{\textit{exp}(x_{i}\cdot y_{i}/\tau)}{\sum_{j=1}^{N}\textit{exp}(x_{j}\cdot y_{i}/\tau)},
\end{equation}
where \(x_{i}\) and \(y_{i}\) are the normalized embedding of the \(i\)-th paired image embedding and text-embedding. \(\tau\) is the temperature term. \(N\) is the total number of samples. Besides, image-text contrastive pretraining with masking strategy \cite{eva,maskclip,scaling,flip} have shown more impressive performance than vallina version, reducing the cost of computational resources. 

\textbf{Masked image modeling (MIM).} Encoders gain enhanced universality and adaptability through the generation or reconstruction of the input\cite{beit,cae,caev2,simmim,mae}, which directs the encoder to learn semantic representations by predicting a percentage of masked regions from the visible ones. The MIM loss is formulated as Eq.~\ref{mim}:

\begin{equation}
\label{mim}
  L_{MIM} = -\mathbb{E}_{t^{m} \sim D} [\log p(t^{m}|t^{v})],
\end{equation}
where the goal is to generate masked image tokens \(t_{m}\) given the visible image tokens \(t_{v}\).
According to the predicted target, MIM can be categorized into pixel-based target and feature-based target:
Early methods, such as MAE\cite{mae} and SimMIM\cite{simmim}, predict the raw pixel values of masked image patches in the input space. While these approaches yield promising results in various downstream tasks, they usually focus on mask strategy and model structure, leading to lower semantic representations\cite{jepa}. Subsequent methods have started to make predictions in representation space like CAE\cite{cae,caev2}, MaskFeat\cite{maskfeat} as an efficient and scalable method for learning semantic representations.

\textbf{Masked language modeling (MLM).} Early masked language models like BERT\cite{bert}, RoBERTa\cite{roberta} are trained to predict masked word tokens \(x_{m}\) given a sequence of tokens \(x_{v}\) also known as "Cloze task", as shown in Eq. ~\ref{eq:mlm}:
\begin{equation}
\label{eq:mlm}
  L_{MLM} = -\mathbb{E}_{x^{m} \sim D} [\log p(x^{m}|x^{v})].
\end{equation}

Furthermore, large language models (LLMs) like GPT-3\cite{gpt-3} predict next token given the prefix language tokens in an autoregressive manner. The pretraining objective is employed in Decoder-only LLMs, as shown in Eq.~\ref{eq:LM}:
\begin{equation}
\label{eq:LM}
  L_{LM}= -\mathbb{E}_{x^{l} \sim D} [\log p(x_{l}|x_{1:l-1})],
\end{equation}
where \(x_{1:l-1}\) denotes the prefix sequence before \(l\)-th token and \(x_{l}\) is the \(l\)-th next token prediction.

\subsubsection{Fine-tuing Techniques}

As foundation models increase in parameter size, they showcase emerging abilities. However, the challenges associated with directly full parameter fine-tuning for new tasks and domains can be a significant concern, including computational resources, complexity, and potential risks such as overfitting. Therefore, to overcome these challenges, we mainly conclude two types of commonly used fine-tune techniques as follows:


\textbf{Prompt tuning} has drawn great attention with the release of GPT-3\cite{gpt-3}. Compared to full parameters fine-tuning, prompt tuning is often more resource-efficient, particularly when computational resources and task-specific data are limited. For example, in-context learning\cite{gpt-3,gpt-4} aims to perform a new task based on few-shot examples provided within the context of a prompt, rather than performing gradient updates through explicit fine-tuning on a large labeled dataset. For hard tasks requiring logic reasoning, such as mathematics, chain-of-thought prompting\cite{chain-of-thoght} provides a detailed, step-by-step explanation or reasoning within the prompt itself, which guides the model to generate more accurate and logical responses.


\textbf{Instruction tuning} is a simple and straightforward multi-task fine-tuning technique that involves adapting a pretrained foundation model to a specific task by providing it with clear, task-related instructions. During inference, the foundation model can generalize to some unseen tasks by explicitly adding task instructions, especially when the inference tasks are similar to those it was trained on. Moreover, it can also help foundation models act as human assistant whose response are better aligned with human intents. Representative works like InstructGPT\cite{instructgpt}, Flan-T5\cite{flan-t5}.

\subsection{Roadmap of Visual Understanding Foundation Models for Road Scenes}
As shown in Fig. \ref{fig:roadmap}, we've recently witnessed a large number of foundation models in road scenes. Here, we divide them into mainly three categories: \textbf{LLM-based foundation model (LLFM)}, \textbf{VLM-based foundation model (VLFM)} and \textbf{vision-only large-vision foundation model (LVFM)} based on the data modality used during pretraining phase.

Kim et al.\cite{kim2018textual} proposed the first framework for generating textual explanations of driving decisions. Despite the novelty of this work, its impact might have been constrained due to reliance on CNN rather than Transformer architecture.

With the prevalence of LLMs in recent years, there emerges numerous foundation models utilizing LLMs (LLFMs) in road scenes. The most popular way to integrate LLMs into the autonomous system is to treat a pretrained LLM as a “brain" that processes structured data from the scene and conducts reasoning. This route prevails after the release of open-source LLMs such as LLaMa\cite{llama}. TrafficGPT\cite{trafficGPT}, DiLu\cite{dilu}, and LINGO-1\cite{lingo1}serialize scene data into structured natural language narrations using task-specific models and feed them into a pretrained LLM for understanding and reasoning. These models just take serialized text data as input, so the performance of these models is highly restricted by the task-specific models, hampering their deployment in real applications.

With the advancements in VLMs, VLFMs seek to train multi-modal encoders for inputs respectively and align them in the manifold space to perform multi-modal understanding tasks in an end-to-end manner. ADAPT\cite{multitask-8} trains a unified end-to-end vision-language Transformer to attend across visual and linguistic input, providing narration, reasoning, and control signal prediction ability. However, ADAPT can only answer rigid-form questions and are unable to perform more complicated tasks such as VQA. The breakthrough turn point comes when multi-modal LLMs like LLaVa\cite{llava} and BLIP2\cite{blip2} release. DriveGPT4\cite{drivegpt4}, Dolphins\cite{dolphins} and many other works\cite{reason2drive, lmdrive, bevclip, detection_6} fine-tune a visual encoder and align the visual features into the language features via Q-former\cite{blip2} to leverage the robust reasoning ability of LLM, contributing to a better human-like understanding of the environment. GAIA-1\cite{gaia1} is a generative world model that synthesizes realistic road scenes from video, text, and action inputs, facilitating a robust model trained on diverse synthetic data and offering fresh opportunities for advancement in intelligent vehicles. Recently, some works\cite{lmdrive, drivemlm} have begun to focus on interactive capability in real-world road scenes, which motivates them to design closed-loop systems that are more realistic for deployment.

In contrast, LVFMs only leverage visual inputs such as camera data and lidar for better feature representation and scene construction, thus facilitating downstream vision-centric tasks without dependence on linguistic data. UniAD\cite{uniad} proposes the first end-to-end framework that incorporates full-stack driving tasks in a unified network. For directly connecting image with lidar in cross-modal localization, LIP-loc\cite{liploc} applies contrastive pretraining to 2D images and 3D points cloud. UniPAD\cite{unipad} proposes a self-supervised framework to enhance feature learning using only images. To exploit the strengths of BEV (Bird's-Eye View) map, BEVFormer\cite{detection_camera_only_1} leverages multi-camera images and spatio-temporal Transformer to construct robust BEV representations. BEVGPT\cite{wang2023bevgpt} integrates prediction, decision-making, and motion planning taking the BEV images as the only input source. As for vision generative models, OccWorld\cite{predict_occ_multimodal_1} is a world model that predicts vehicle and scene dynamics in a 3D occupancy-based representation.

\section{Common Practices on Visual Understanding Models for Road Scenes}
\label{sec:common_practices}

\begin{figure*}[t]
    \centering
    \includegraphics[width=\linewidth]{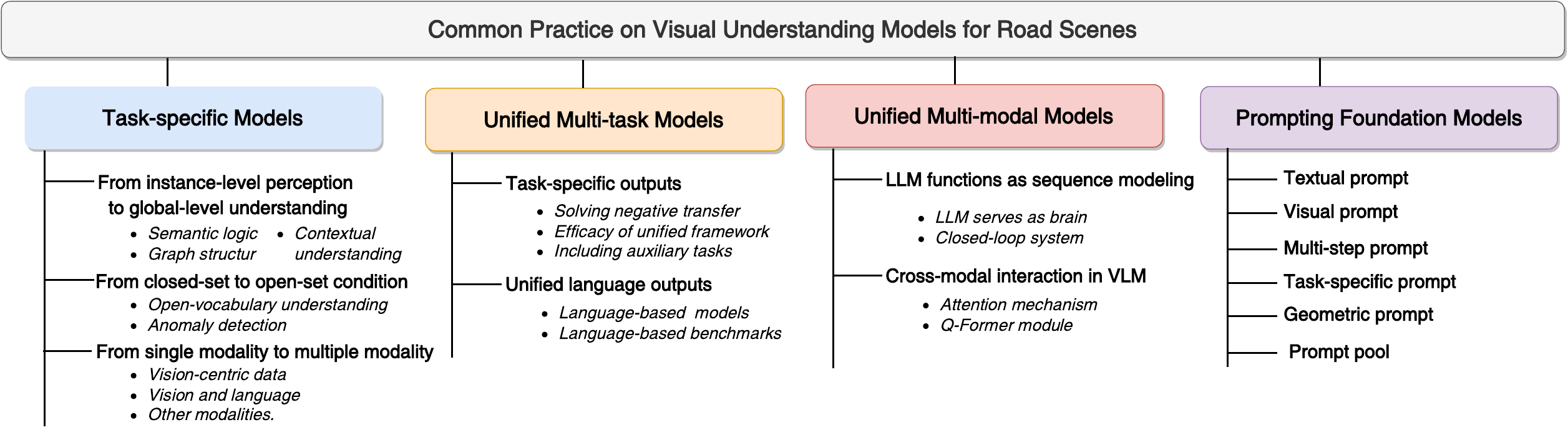}
    \caption{The overall framework of Sec.~\ref{sec:common_practices}. We review existing works on visual understanding for road scenes from four perspectives: task-specific models, unified multi-task models, unified multi-modal models, and prompting foundation models, respectively. By organizing the review based on these four perspectives, the framework provides a structured and insightful exploration of the existing literature on visual understanding for road scenes.}
    \label{fig:common_practice}
\end{figure*}
In this section, we provide a comprehensive review of existing works on road scene understanding from four perspectives, as illustrated in Fig. ~\ref{fig:common_practice}. We first introduce task-specific models in Subsec.~\ref{sec:task-specific}, unified multi-task models in Subsec.\ref{sec:unified_multi-task}, unified multi-modal models in Subsec.~\ref{sec:unified_multi-modal}, and prompting foundation models in Subsec.~\ref{sec:prompting}, respectively.
\subsection{Task-specific Models}
\label{sec:task-specific}

Task-specific models for road scenes are designed for specific downstream tasks. These models are trained on specific data and architecture to optimize their performance in those particular tasks, making them to be highly proficient in their designated areas. In this subsection, we conclude them from three perspectives: from instance-level perception to global-level understanding, from closed-set to open-set condition, and from single-modality to multi-modality. 

\textbf{From instance-level perception to global-level understanding.} Unlike the conventional approach of recognizing individual traffic participants\cite{vanishing,camera,3dlane}, modern methods prioritize analyzing the environment through a broader, global-level understanding.
For example, semantic logic between instances serves as global-level comprehension. Understanding overall information from traffic signs solely based on instance-level perception can be challenging. Yang et al. \cite{detection_0} introduce a novel traffic sign interpretation task that involves both localizing and recognizing traffic signs, aiming to provide a precise global semantic understanding akin to natural language for road scenes. This approach makes a great innovation by emphasizing the consideration of broader relationships among individual traffic signs and generating global accurate traffic instruction information. Similarly, contextual understanding\cite{contextual,seg_2} excels in achieving a comprehensive global perspective. Mask2Anomaly\cite{seg_2} casts traditional road anomaly segmentation task as a context-aware mask classification problem by considering contextual semantics around the anomalies. Compared with pixel-based architecture which densely predicts label for each pixel, mask-based Transformer architecture is more advantageous for anomaly segmentation. This is attributed to the mask-based architecture promoting objectness, facilitating the comprehensive capture of anomalies as complete entities. This approach yields more consistent anomaly scores and diminishes false positives.
Graph representation is another kind of global understanding. GP-Graph \cite{predict_pedestrain_multimodal_1}reconceptualizes the multi-modal pedestrian trajectory prediction task by incorporating both inter- and intra-group relations using graph representations. Unlike existing interaction models treating each pedestrian as a graph node, GP-Graph dynamically segregates pedestrians in crowded settings into individual groups. This innovation enables effective attention to both inter-group and intra-group interactions, significantly enhancing prediction accuracy and facilitating model interpretation. 
Unlike instance-level perception which focuses on individual objects without exploiting relationships between objects, global-level understanding lies in its ability to provide a holistic understanding for road scenes, by considering the interactions and dependencies across various objects.

\textbf{From closed-set to open-set condition.}
Previous detection methods\cite{shi2023cobev,haar2024measuring} and segmentation methods \cite{ground,3mt}for road scenes primarily aimed to achieve high performance within common class sets such as pedestrian and car, often overlooking considerations for underlying unseen objects in open-set scenarios, such as a dog on the highway. 

Open-set understanding in autonomous driving refers to long-tailed corner cases detection\cite{li2022coda}, which holds significant importance in road scenes, particularly for ensuring safety and reliability. Previous method \cite{synthesis} has endeavored to generate corner samples to maximize the coverage with pretrained models, which is cost and impractical. However, a growing trend involves discovering unseen objects through the utilization of external knowledge sources, such as multi-modal data and pretrained vision-language models. As an example, SalienDet \cite{saliendet} solves open-world detection in road scenes via salience maps-based feature enhancement techniques for unknown objects. Likewise, GOOD \cite{good} delves into extracting geometric cues of traffic objects from depth and normal maps within an open-world context, demonstrating that geometric information is a crucial component for open-set understanding. On the other hand, open-vocabulary understanding targets the recognition of arbitrary objects based on textual input by leveraging advancements in vision-language learning. OVTrack \cite{ovtrack} extends a closed-set object tracker into an open-vocabulary tracker by leveraging knowledge distillation from a pretrained vision-language model \cite{clip}. Additionally, it also mitigates the scarcity of road data through a strategy of hallucination-based generation harnessing the power of the diffusion-based model. Anomaly detection also represents an open-set scenario. Bogdoll et al. \cite{detection_5} delve into the challenge of unexpected situations within road scenes. They explore the application of world models for the demanding anomaly detection task and showcase their seamless integration with current approaches. Having the ability to appropriately recognize inputs that belong to classes not encountered during the training phase, the open-set capability is a crucial aspect for ensuring the safety of intelligent vehicles in road scenes.

\textbf{From single modality to multiple modalities.} Unlike previous methods have the limitation in using only single modality\cite{mixsup,lidarptq,realtime}, the utilization of multiple modalities proves beneficial in obtaining a comprehensive understanding of road scenes\cite{multimodalfusion1,multimodalFusiontransformer,imagebind}. 

Vision-centric data. Many methods have shifted their attention to multiple visual sensor data, ranging from camera-only\cite{instanceaware,panoocc,multicamera,detection_camera_only_1} 
or lidar-only\cite{idet3d,offroad,detection_lidar_only_1,opensight,frnet}, to collaborate vision-centric sensors\cite{fusion,voxelnextfusion,sparsefusion,transfuser}. These methods for collaborating sensors mainly involve two categories, multi-sensor fusion and multi-sensor calibration. Multi-sensor fusion methods\cite{fusion,voxelnextfusion,detection_1,detection_3} aim to observe road scenes via utilizing complementarity across sensors in a fusion fashion. TransFusion \cite{fusion} introduces an enhanced LiDAR-Camera fusion approach utilizing a soft-association mechanism for 3D Object Detection. This method is specifically designed to tackle challenges arising from suboptimal image conditions, such as intensive illumination, thereby promoting greater robustness in fusion solutions. During actual driving, multiple sensors on the ego-vehicle may experience positional displacement and misalignment issues. To address these issues, multi-sensor calibration\cite{seg_1,robust,soac,calibformer,instanceaware} is proposed. Jiang et al.\cite{seg_1} not only use rotational angle noise data augmentation to simulate the poor calibration of multi-modal information but also introduce knowledge distillation strategy to prevent the model from experiencing performance degradation due to noise data augmentation when the model is well-calibrated.

Vision and language. Recent progressions have extended the scope of basic visual modalities from sensor combinations to encompass the fusion of vision and language modalities. Present methodologies\cite{vlpd,language-guided,incorporating,large,3dunsupervised} focus on further integrating linguistic comprehension into vision-centric tasks. VLPD\cite{vlpd} introduces a novel pedestrian detection approach aimed at overcoming challenges posed by confusion, small scales, and occlusion. This method self-supervises its own detector using segmentation pseudo-labels generated by pretrained vision-language model and learns discriminative pedestrian features through contrastive learning.

Other modalities. For instance, within the realm of risk assessment, driver emotions\cite{predict_cpsor,defining} also serve as valuable information sources. CPSOR-GCN\cite{predict_cpsor} have shifted their attention to exploring multi-modal training strategies on enhancing performance in trajectory prediction. Besides considering the physical features in road scenes, their research proposes a multi-modal prediction framework incorporated with cognitive theory\cite{cognitive_theory}, simultaneously considering the environment stimuli, emotion state, and behavior of the driver.

\subsection{Unified Multi-task Models}

\label{sec:unified_multi-task}
\begin{figure}[t]
    \centering
    \includegraphics[width=\linewidth]{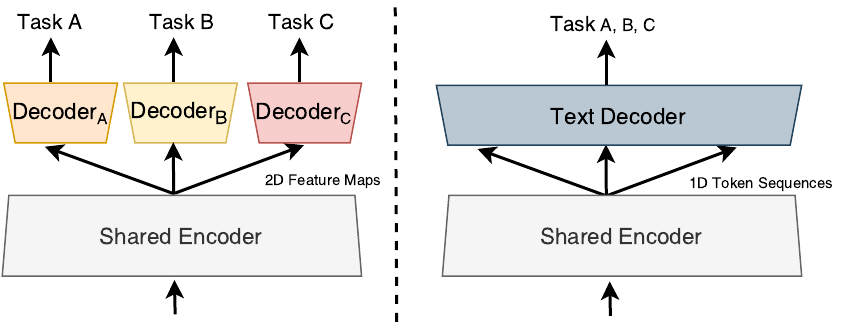}
    \caption{Unified multi-task models can be categorized based on their outputs into two distinct types. The first type (left) includes models with task-specific outputs, characterized by a shared encoder and individual task-specific heads across all tasks. In this type, the shared encoder processes the input data to produce 2D feature maps, and each task has its dedicated head to generate task-specific output, respectively. Conversely, the second type (right) refers to models with unified language outputs. These models consist of a shared encoder and a unified text decoder to generate texts for all tasks. The shared encoder is responsible for transforming the input data into 1D token sequences, contributing to language-based representations for all tasks.}
    \label{fig:unified_multi-task_models}
\end{figure}

Understanding road environments is an essential requirement for ensuring safe autonomous driving. In the past few years, although many methods have been proposed to solve a single task with satisfactory performance, there are still many difficulties in the synergy of various tasks across granularities.
Multi-task learning (MTL) aims to jointly train on the unified models through shared parameters across various tasks, enabling parameter efficiency and better performance compared with single-task learning. Recently, researchers have begun to explore integrating various visual understanding tasks in a unified framework\cite{multitask-00,multitask-01}. We classify these methods into two categories according to multi-task output formats, as illustrated in Fig.~\ref{fig:unified_multi-task_models}.

\textbf{Task-specific outputs.} These methods, usually consisting of one encoder for shared features and task-specific heads, have been adopted in a wide range of tasks such as 3D object detection\cite{detection_1,detection_2}, RGB-X object detection\cite{detection_3}. 

Solving the negative transfer problem across tasks is the primary concern for MTL. In this trend, VE-Prompt\cite{multitask-1} aims to generate task-specific features for different task via the prompting of visual exemplars. This method provides novel insights into the effect of task-specific prototype-based prompts on vehicle detection, lane detection, and drivable area segmentation tasks. It uses CLIP as a visual prompt generator to generate learnable task-specific prototypes from visual exemplars, and then these task-specific prototypes are attended with features extracted by one shared backbone to acquire high-quality task-specific features for decoders via cross-attention mechanism. In pursuit of effective information propagation across diverse tasks, CML-MOTS\cite{cmlmots} introduces a novel multi-task framework designed for video instance segmentation and tracking. This framework incorporates an innovative associative connection across different task heads, facilitating the fusion of outputs from various task heads.

The efficacy of a multi-task model stems from its ability for tasks to mutually benefit from each other. Therefore, real-time unified perception and fast inference are critical considerations. For example, YOLOP\cite{multitask-4} extends real-time and lightweight objector YOLOv5 to multi-task model for performing traffic object detection, drivable area segmentation, and lane detection simultaneously. With the same goal of fast inference, YOLOM \cite{look} introduces a real-time and lightweight multi-task model designed for joint object detection, drivable area segmentation, and lane line segmentation tasks. The model enhances generalization without the need for customizable design by incorporating a unified loss function for all segmentation tasks. Additionally, the adoption of a series of convolutional layers as the segmentation head significantly reduces the overall inference time. LiDAR-BEVMTN\cite{lidarbevmtn} proposes a LiDAR-only multi-task perception system for collaborative detection, semantics, and motion segmentation. Opting for LiDAR-only perception in Bird's-Eye View (BEV) provides an alternative for fast inference on embedded devices. This consideration arises from the signal sparsity of radar sensors, which operate at long ranges and may exhibit lower quality. Moreover, dense data from camera sensors, which are susceptible to diverse weather conditions, further underscores the appeal of this approach for efficient inference.  

Including auxiliary tasks to enhance the performance of the main task is another key advantage of MTL. The concept of world models\cite{world_model} has been widely discussed recently with the following characteristics of remembering history, learning experience, modeling the world, and predicting the future. For example, OccWorld\cite{predict_occ_multimodal_1}is a world model in predicting future scenes with the auxiliary ego-car motion trajectory prediction. Through efficiently discretely encoding \cite{vqvae} to obtain high-level semantic representations of past frames and a spatial-temporal autoregressive-based generative transformer to enable interaction world tokens in the same and cross timestamp, task-specific decoders can make a more reasonable prediction than ground truth labels generated by self-supervised methods.

\textbf{Unified language outputs.} While it is a common practice for multi-task models to produce task-specific formats as results, certain limitations persist. Vision-centric tasks often exhibit distinct output formats, varying across different axes such as granularities and the number of classes. A novel approach to address the issue is to use natural language as the unified output. 
For example, HiLM-D\cite{multitask-6}is a LLM-augmented model that it performs risk object localization and intention and suggestion prediction in a natural language manner, without exquisite architecture design. ADAPT\cite{multitask-8} jointly predicts the vehicle's action and gives the reasons based on the observation of surrounding scenes in the caption manner. Furthermore, traffic sign interpretation is a new task proposed by\cite{multitask-5}, which globally detects, and recognizes traffic signs, and gives accurate traffic instruction information in logic-based natural language description. 

Newly proposed prompt-based benchmarks provide a convenient way to build unified multi-task models using natural language as unified multi-task outputs. Specifically, NuPrompt \cite{languageprompt} expands the Nuscenes dataset\cite{nuScenes} to instance-text pair by constructing language descriptions for video clips. Similarly, NuScenes-MQA \cite{NuScenes-MQA} provides an evaluation of a model's capabilities in sentence generation and VQA annotated in a questions and answers (QA) format.  DriveLM \cite{drivelm} also facilitates perception, prediction, and planning with logical reasoning. DRAMA\cite{multitask-7} aims to simultaneously detect risk objects and give explanations.

\subsection{Unified Multi-modal Models}

\begin{figure}[t]
    \centering
    \includegraphics[width=\linewidth]{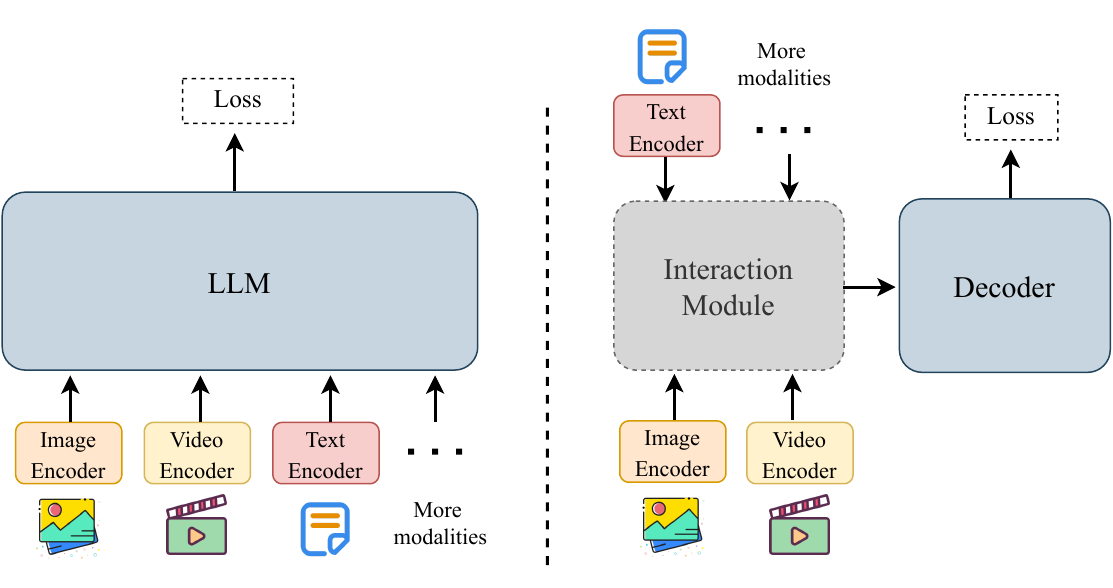}
    \caption{Comparison of LLM-based (left) and VLM-based (right) unified multi-modal models. The LLM-based model takes LLM as a center place, which transforms multi-modal data into textual tokens that are easily modeled by LLM in the manner of sequence modeling. The VLM-based model emphasizes cross-modal interaction involving fusion, alignment, and matching across multi-modal data.}
    \label{fig:unified_multimodal_models}
\end{figure}

\label{sec:unified_multi-modal}

In recent years, the advancements in natural language processing through LLM and the progress in seamlessly integrating visual and linguistic comprehension via VLM, have brought a transformative paradigm shift in road scene understanding. This evolution has given rise to a pursuit of a more comprehensive and unified approach to multi-modal understanding. LLM-based models, particularly multi-modal LLM (MLLM), specialize in language-related generation tasks. Moreover, owing to the robust understanding and reasoning capabilities of LLMs, these models\cite{lmdrive,drivegpt4} are trained to learn patterns and semantics from a broader spectrum of modalities beyond vision and language, such as control signs of the intelligent vehicles. On the other hand, VLM-based models\cite{dolphins} are designed to comprehend visual and textual data in pairs and bridge the gap between images or videos and the corresponding texts. They emphasize the intricate matching and interaction between vision and language. Comparison of LLM-based and VLM-based multi-modal models is illustrated in Fig.~\ref{fig:unified_multimodal_models}. In this subsection, we discuss their specific strengths on contributing to unified multi-modal scene understanding as follows:

\textbf{LLM functions as sequence modeling.} In contrast to module design in previous driving systems, introducing LLM as the central brain for various tasks has become a prominent fashion. This kind of methods usually consists of modality-specific encoders for multi-modal input and one unified MLLM as a decoder for various tasks. They focus on transforming the driving environment into serialized textual tokens which can be easily modeled by Transformer architecture. For example, given the poor generalization of previous methods in the open-world environment, GPT-Driver\cite{gptdriver} treats LLM as a reliable motion planner for autonomous vehicles by reformulating the motion planning task as a language modeling problem. Specifically, it converts perception and prediction results into language descriptions which are viewed as input prompts for LLM and then decoded to generate ego-movement trajectory in language-format outputs. 
DriveGPT4\cite{drivegpt4} not only utilizes a video tokenizer to encode video frames and a shared text tokenizer to tokenize user query and past control signals, but employs LLM to predict desired answers and future control signals. LanguageMPC\cite{languagempc} harnesses the reasoning capability of LLM to analyze complex road scenes and make high-level textual decisions. These textual decisions are then converted into mathematical representations to seamlessly guide the bottom-level controller for Model Predictive Control. Similarly, some related works have been proposed in open-loop settings(such as Driving-LLM\cite{drivingwithllm}), where the output or behavior of the model is not directly considered or influenced for real-time feedback.

However, these methods discussed above ignore real-time feedback from the environment in an open-loop setting. Closed-loop driving systems are essential for safe and adaptive driving because they allow vehicles to respond to changes in real time. Typically, \cite{drivelikehuman}, serving as a catalyst, demonstrates impressive abilities in real-time interaction with the environment, human-like driving with common sense, and memorization to overcome catastrophic forgetting, especially when faced with long-tailed corner cases.
LMDrive\cite{lmdrive} is a novel language-based closed-loop autonomous driving framework. LMDrive integrates all visual tokens of past timestamps, language tokens tokenized from multi-modal multi-view visual data and language instructions, respectively. Control signs are then predicted until the given instructions are completely conducted.  DriverMLM\cite{drivemlm} serves as MLLM-based intelligent agents. In addition to the commonly mentioned visual signals, it also employs interfaces for driving rules and user commands to build a comprehensive understanding of road scenes and align the output of MLLM with behavioral planning states predicted by the off-the-shelf behavior planning module.
Wang et al.\cite{drivesafe} introduce LLM as a behavior planer with safety assurance. The LLM-based driver agent simultaneously ensures safety and reinforces performance in driving tasks, which takes intention prediction, scenario description, behavior state, its own memory of past scenarios, and experience as input, and outputs safety-constrained behavior decisions. Moreover, rendered new observations from the environment are provided and next actions are made based on new observations iteratively. 
More uniquely, in contrast to directly tokenizing environment observation into language tokens, Agent-Driver\cite{languageagent} designs a tool library to abstract them via function call. The text-based messages are returned, leading to a precise and intensive environment description. Furthermore, cognitive memory for storing common sense and past experience serves as complementary knowledge to enhance the reasoning ability of MLLM to output safe and comfortable actions.

\textbf{Cross-modal interaction in VLM.} Methods\cite{tian2024drivevlm} in the VLM-based unified multi-modal models differ from the ones discussed earlier by incorporating an interaction module for multi-modal tokens before processing them with the decoder. Within this interaction module, multi-modal tokens undergo fusion, alignment, or matching processes.
Dolphins \cite{dolphins}, based on openFlamingo\cite{openflamingo} which features in-context learning capabilities, inserts new gated cross-attention layers within a pretrained frozen LLM with the LoRA\cite{lora} module. These newly added layers promote stable interaction between vision and language via gate-mechanism. To improve interpretability during the decision-making process, Reason2Drive\cite{reason2drive} views the decision-making process as chain-based reasoning task. The algorithm proposes a combination of vision encoder and prior tokenizer, embedding multi-modal data into latent space, respectively. A Q-Former serves as an alignment module followed by an LLM and vision decoder responsible for predicted answers to user questions and perception results.
Drive-Anywhere\cite{driveanywhere} uses multi-modal foundation models to extract patch-level features rather than vector representation for the entire image. To improve generalization to new driving conditions, latent space simulations enhanced language modality is conducted by replacing original visual features with substitute contextual concepts.
Talk2BEV\cite{talk2bev} utilizes MLLMs to acquire captions for each object in the Bird's Eye View (BEV) maps, as well as descriptions for the overall scene. By aligning this semantic information with the inherent spatial information in BEV maps, metadata is generated, i.e., the LLM-enhanced BEV map. The understanding of the current BEV map is then achieved by prompting the LLM using a Chain of Thought (CoT) approach.
PromptTrack\cite{languageprompt} incorporates a novel prompt reasoning branch based on current object tracking algorithms to deal with referring object tracking whose objective is to track desired objects according to the user's prompt. Specifically, a cross-modal interaction module is implemented to acquire prompt-aware features and the refined features are further decoded by the tracking head.

\subsection{Prompting Foundation Models}

\label{sec:prompting}
The emergence of the multi-modal prompts have indeed triggered a paradigm shift for transfer learning. Currently, an emerging trend is to highlight the significance of prompt engineering on transfer learning. For example, Liang,et al.\cite{effective} proposes an effective pretrain-adapt-finetune paradigm for unified multi-task learning. In the phase of adapt stage, task-specific concepts are utilized as prompts to maintain consistency between visual features and textual priors, overcoming negative transfer within multiple tasks. Xu, et al.\cite{pretrainpromt} focus on time-to-event analysis to ensure safety in cyber-physical system (CPS), e.g., autonomous driving systems. To alleviate data scarcity, they adopt a novel transfer learning method, namely "pretrain-and-prompt tuning". Specifically, firstly pretraining on large-scale datasets to build fundamental knowledge, and then transfering source knowledge to the target CPS via prompt tuning. In the phase of prompt tuning, it consists of three steps: prompt template designing, answer generation and answer mapping. 

Multi-modal prompts play a crucial role in guiding a model's understanding and reasoning capabilities. These prompts leverage diverse types of information or cues, such as images, text, or other modalities, to enhance the model's comprehension of the environment and improve its decision-making processes. We summarize foundation models with various multi-modal prompts as follows:

\textbf{Textual prompt.}
In the context of driving-related models, textual promptable models like DriveMLM\cite{drivemlm}, GPTDriver\cite{gptdriver} and DriveGPT4\cite{drivegpt4}, refer to the integration of natural language where textual instructions or commands are used to direct or interact with the autonomous vehicle (AV). These prompts serve as a form of human-vehicle communication, and can guide the AV's actions or decision-making processes.

\textbf{Visual prompt.}
Visual promptable models\cite{gpt-4v,Talk2Car,talk2bev} involve utilizing visual cues to enhance the scenario understanding. The representative algorithms in this line include multi-view vision-prompt\cite{multiviewvisionpromptfusion}, cross-modal prompt fusion\cite{emotionrecognition}, and learnable visual prompts\cite{multitask-1} in various driving scenarios.
As the road scene becomes increasingly complex, the plain textual and visual promptable models no longer meet the demands. They demands real-time responses and effective adaption to new scenarios. Novel prompts beyond simple vision and language cues are being explored to enhance autonomous systems and their capabilities.

\textbf{Multi-step prompt.} Existing prompt techniques mentioned above solely activate the powerful reasoning ability of FMs via a single step, where it is sub-optimal to exploit powerful world knowledge from LLM. 
Typically, CoT embodies the idea that language is not just a collection of individual words or sentences but rather multi-step connected thoughts or concepts. Observation of the lack of fine-grained understanding and reasoning ability in existing VLMs, Dolphin\cite{dolphins} extends generic image datasets to instruction-following datasets with grounded CoT response, enriched with detailed multi-step reasoning. The model is then trained on the enriched image instruction-following datasets to establish more general understanding and reasoning ability. To transfer the ability learned from generic images to the driving domain, the process also refers to finetuning on driving-related video instruction-following datasets.
GPT-driver\cite{gptdriver} argues that a sequential prompting-reasoning-finetuning strategy is effective for highly encouraging reasoning ability of FMs. This strategy involves a first stage for prompting LLM observed on perception and prediction, a second stage for multi-step CoT reasoning, and the last finetuning stage for aligning LLM's outputs with human driving trajectories.
Moreover, OpenAnnotate3D\cite{zhou2023openannotate3d} introduces an open-vocabulary auto-labeling pipeline to generate 2D/3D mask and 3D bounding boxes for multi-modal data, which greatly reduce the labor of extensive manual labeling. The algorithms can understand given target description with the CoT reasoning of LLM, and then provide high-quality annotations with multi-modal alignment.

\textbf{Task-specific prompt.} Multi-task learning involves joint training to perform multiple tasks simultaneously, leveraging shared representations to improve overall performance. However, those prompts learned from a specific task might struggle to transfer effectively to entirely different tasks in a multi-task framework. This limitation is mainly due to the specificity and task-oriented nature of the prompts. Consequently, The use of task-specific prompts helps guide the model's learning process for individual tasks, preventing negative transfer within tasks. For example, Liang et al. sequentially proposes two methods\cite{effective,multitask-1} to effectively solve gradient conflict in multi-task learning from the perspective of generating task-specific features. They both treat CLIP model as a task-specific prompt generator, but the difference is that task-specific prompts are generated from language concepts and visual exemplar individually. 
\cite{effective} firstly constructs a task-specific sentence concatenated by class names and the combination of textual embeddings that are tokenized from the sentence via the text encoder of CLIP, and some learnable tokens are viewed as task-specific prompts.
However, for each task, VE-Prompt\cite{multitask-1} firstly crop class-related regions and 
the image encoder of CLIP is adopted to extract features, and the average of these features is used as task-specific prompts.
TaskPrompter\cite{taskprompter2023} is a novel multi-task prompting framework with joint 2D-3D scenarios understanding, referring to 3D vehicle detection, semantic segmentation, and monocular depth estimation. With the help of task prompt embedding, it aims to unify learning of task-specific and task-generic representation as well as cross-task interaction in a compact model capacity, rather than separately learning in specific modules. In this way, the task prompts are mapped into spatial- and channel-prompts to promote interaction patch tokens from input image in the spatial and channel dimensions, and further serve as prompting dense task-specific features and multi-task outputs. 

\textbf{Geometric prompt.} Attention mechanisms have been incredibly useful in sequence modeling. However, they do face limitations in quadratic-scaled computational complexity and memory constraints, particularly when dealing with very long sequences. Researchers are actively exploring various methods to mitigate these limitations with the help of a prompt-based attention mechanism. For example, to address the challenge of modeling human state change given long-range series signals, Niu et al.\cite{windowprompt} incorporate \textit{window prompt} to enable flexible and efficient attention on a local window. Similarly, DriveGPT4\cite{drivegpt4} points out that a vector representation for the entire image significantly causes the loss of spatial information, which is crucial for understanding structural road scenes. Therefore, it innovatively replaces the vector-wise with patch-wise representation extracted by multi-modal FMs. This approach involves constructing attention masks via \textit{anchor prompt}, allowing attention modules to pay attention to specific regions rather than global regions.

\textbf{Prompt pool.} In continual learning, designing the prompt pool that stores previous task-specific prompts is a effective way to overcome catastrophic forgetting, allowing the model to improve its performance over time. I3DOD\cite{i3dod} points out that current incremental 3D detectors fail to model the relationship between object localization and semantic labels and then introduce task-shared prompts maintained by a prompt pool to learn localization-category matching.

In summary,  multi-modal prompts play a crucial role in enhancing the capabilities of rich information fusion, contextual understanding, and adaptability to diverse environments by providing a richer, more diverse set of cues.

\begin{figure*}[t]
    \centering
    \includegraphics[width=0.9\linewidth]{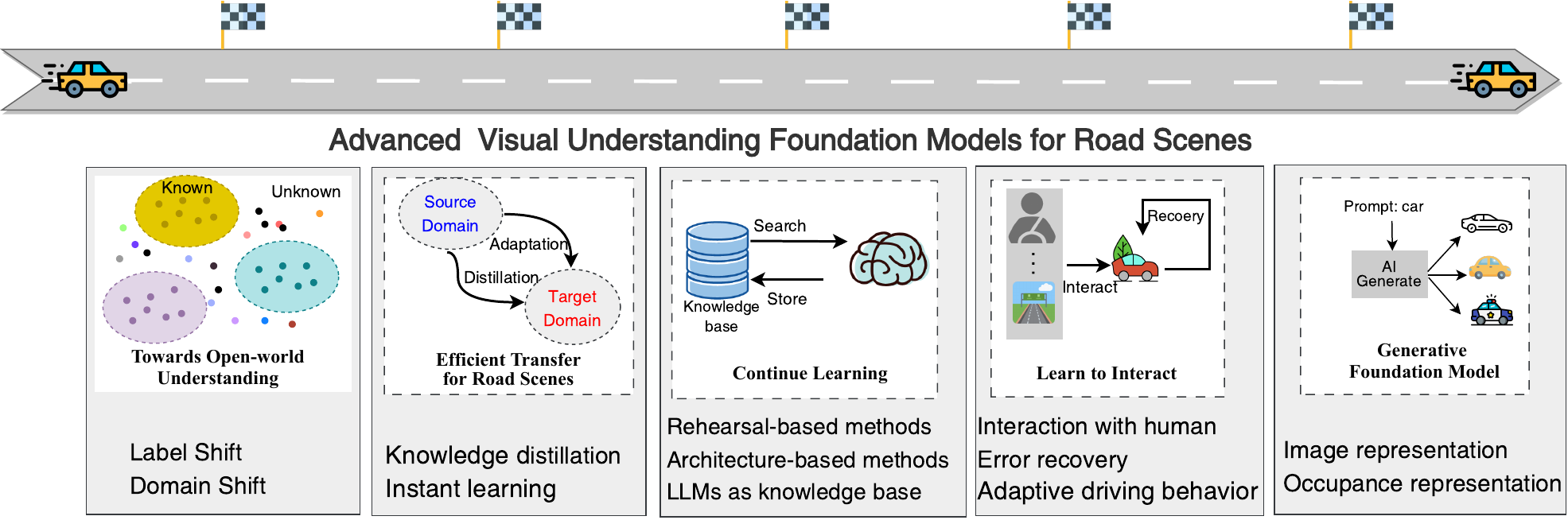}
    \caption{The overall framework of Sec.~\ref{sec:advanced}. The advanced capabilities of visual understanding foundation models are highlighted from five perspectives: towards open-world understanding, efficient transfer for road scenes, continual learning, learn to interact, and generative foundation models, respectively.}
    \label{fig:advanced_model}
\end{figure*}

\begin{figure*}[t]
    \centering
    \includegraphics[width=0.9\linewidth]{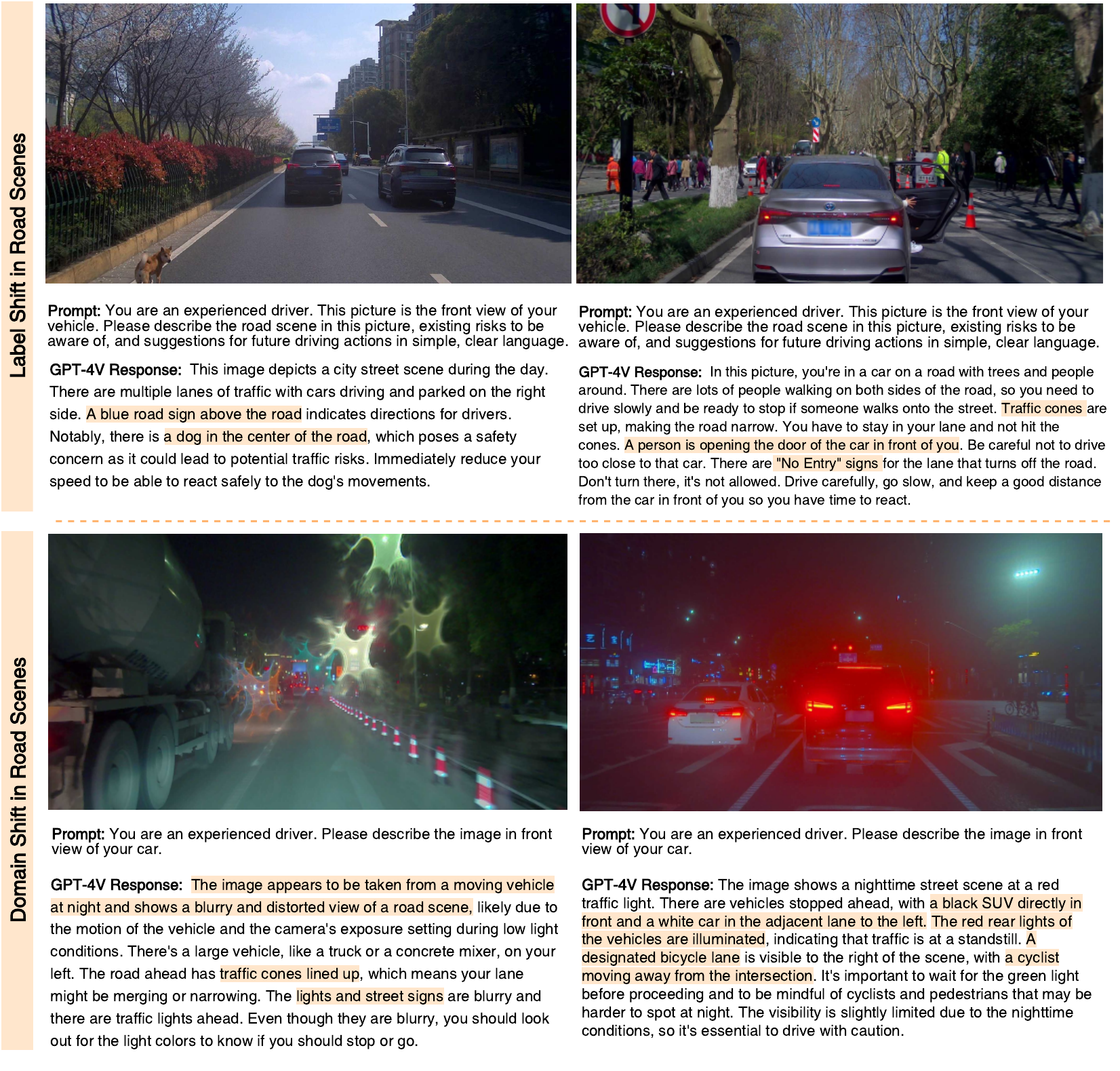}
    \caption{Illustration of GPT-4V's capability on open-world road scene understanding in both label shift and domain shift scenarios. The example images are from CODA\cite{li2022coda}. Given the prompts, right answers in GPT-4V's response are highlighted. These results reveal that GPT-4V can identify scene objects and further provide advice for safe driving, underscoring remarkable understanding capability of existing foundation models.}
    \label{fig:open-world}
\end{figure*}
\section{Advanced Visual Understanding Foundation Models for Road Scenes}
\label{sec:advanced}

In this section, we aim to highlight the advanced strengths of MM-VUFMs on diverse learning paradigms, as illustrated in Fig.~\ref{fig:advanced_model}. These strengths demonstrate the versatility of MM-VUFMs in addressing complex conditions that frequently occur in road scenes.
\subsection{Towards Open-world Understanding}
The real-world driving scenario is characterized by its dynamic and complex nature, constituting an open-world environment that challenges autonomous vehicles. Unlike controlled and static settings, the landscape of the road is constantly changing, presenting a multitude of variables and unpredictable elements. Existing foundation models have made great progress towards open-world understanding. We show GPT-4V's\cite{gpt4v-1,gpt4v-2} understanding capability on open-world road scenes with both label shift and domain shift in Fig.~\ref{fig:open-world}, considering that the access to other models was not released. 

\textbf{Label shift} refers to great changes in label distribution between training data and real-world scenarios encountered during deployment. Existing methods often lack the ability to fully understand long-tailed road scenes, such as animals on the road and overturned trucks. These extreme conditions underscore the necessity for further advancements in open-world generalization. In this direction, DriveLM\cite{drivelm} significantly improves generalization to novel objects encountered during inference by introducing a graph structure, where the decision-making process is regarded as a graph-based reasoning task. In this structure, each vertex represents the VQ pair relevant to key objects in the current scene, and logical dependencies between adjacent vertices are reformulated as edges. Unlike previous methods that regard question-answer (QA) pairs as independent individuals, this graph of QA pairs with logic dependencies enables logical reasoning and offers more promising zero-shot generalization. To explore the effect of semantic information on detecting unseen objects, Elhafsi et al. \cite{semantic} adapt an LLM endowed with contextual understanding and reasoning capabilities to identify semantic anomalies. In the proposed framework, the current environment's observation is translated into a natural language description. This scene description is subsequently incorporated into a prompt template and sent to the LLM, which utilizes contextual reasoning to identify potential semantic anomalies in the described scene. 
Moreover, existing datasets tailored for road scenes usually fall short of adequately covering unseen cases due to the substantial label cost. This limitation results in a lack of exposure to rare situations and poor generalization to unknown objects. To address this challenge, data-driven methods are emerging. For example, TrafficSim \cite{trafficsim} proposes a solution by generating large-scale data through simulation. This simulation is constructed and learned directly from real-world data, incorporating human intentions to ensure a more comprehensive representation of diverse road scenes. Similarly, KING\cite{king} proposes a novel gradient-based safety-critical road scene generation in the CARLA simulator.

\textbf{Domain shift} occurs when there is a misalignment between the source domain where the model is trained and the target domain where the model is applied. For road scenes, the domain shift implies that the model may encounter environments, scenarios, or conditions during deployment that differ from what it experienced during training. This can lead to a decrease in performance as the model struggles to generalize effectively to unseen domains.
Wen et al.\cite{gpt-4v}has undergone extensive road experiments on real-world road scenes. Observations indicate that GPT-4V has demonstrated superior understanding and inference capabilities in dealing with unseen scenarios compared to existing driving agents.
DriveAnywhere\cite{driveanywhere} innovatively alleviates the domain shift problem on unseen scenarios via simulating different scenarios in latent space. 
Specifically, a series of concepts relevant to road scenes are obtained from the LLM. The textual features of these concepts are computed and used to replace visual features, simulating various scenarios based on the principle of exceeding the similarity threshold. This simulation technique also acts as a data augmentation strategy in the latent space, exposing models to diverse scenarios during training. Ultimately, this process enhances the model's ability to generalize in open-world environments during deployment. Inspired by VLM with rich knowledge, Dolphins \cite{dolphins} is built upon a general vision-language model\cite{openflamingo}, and leverages the underlying knowledge gained from pretraining. This approach aims to enhance in-context learning to improve generalization across a spectrum of driving-related tasks via few-shot scene images.

In summary, effectively addressing label shift and domain shift is essential for ensuring the robust performance of intelligent vehicles in an open-world environment. By mitigating the challenges posed by shifts in label distribution and domain conditions, the model can enhance their adaptability, generalize effectively, and navigate the dynamic complexities of real-world road scenes with better reliability and safety.

\subsection{Efficient Transfer for Road Scenes}
\label{section:transfer_ability}

Efficient transfer for road scenes involves adapting generic knowledge learned from natural scenes to road scenes. Due to a great domain gap between natural scenes and road scenes, this means that foundation models(FM) are required to not only distillate foundational knowledge gained from the pretraining stage to achieve good generalization on common objects (e.g., car, pedestrian), but also to be applicable to specific policies which are unique in road scenes.
Moreover, instant learning further enhances the adaptability of foundation models. In the dynamic landscape of the road, where conditions can change rapidly, the model continuously updates its understanding. This process allows it to adapt to evolving situations, making informed decisions in the face of dynamic changes on the road. Therefore, we discuss knowledge distillation and instant learning for efficiently transferring foundational knowledge of FMs to new road scenes.

\textbf{Knowledge distillation} refers to the process of transferring foundational knowledge from large and general FMs \cite{clip,sam,chatgpt} to small and specialized models. FMs have learned foundational understanding, reasoning, and decision-making capabilities from widespread data and scenes. The small and specialized model is a lightweight model tailored specifically for driving-related tasks. However, directly deploying a large model onto autonomous vehicles might be computationally expensive and impractical due to the constraints in memory, power. To address this, knowledge distillation allows for the compression and transfer of the FM's knowledge into specialized models for downstream tasks. 

Distill CLIP knowledge to road scenes. With the remarkable success of multi-modal contrastive pretraining, the recognition of its pivotal role in cross-modal localization tasks has come to the center. LIP-Loc \cite{liploc} is the first work to solve global visual localization utilizing image-LiDAR contrastive pretraining. It aims to learn a joint embedding space for image and LiDAR using the same contrastive symmetric loss in CLIP and achieve zero-shot pose prediction about the given image within expansive point clouds.  LiDAR-based retrieval has also emerged in LidarCLIP\cite{lidarclip}. Directly aligning the features of point clouds with text features is a non-trivial solution due to the scarcity of LiDAR-text data. Instead, image features serve as a bridge for connecting text and lidar data. For each training pair of image and lidar, lidar data are first transformed into image planes and then fed to a lidar encoder to fit the features of a frozen CLIP image encoder by maximizing the similarity. In 3D scene understanding, CLIP2Scene\cite{clip2scene} leverages CLIP knowledge to solve point cloud semantic segmentation via semantic and spatial-temporal consistency regularization.

Distill SAM knowledge to road scenes. Intermediate representations generated by FMs on the fly provide a rich source of knowledge, contributing to the adaptability and effectiveness of various tasks. In multi-modal 3D object detection, segment anything (SAM) has raised great attention from the research community. Chen et al. propose an extension of SAM\cite{sam} to the 3D task using VoxelNeXt\cite{voxenext}, a fully sparse 3D object detector. With prompts in the form of points or boxes, the model outputs a combination of a 2D mask and a 3D bounding box. The incorporation of promptable SAM significantly reduces the annotation cost for 3D detection. RoboFusion \cite{robofusion} is a robust framework to suppress the noise in the driving scenes via SAM. To adapt SAM to road scenes with substantial noise, SAM-AD is obtained via mask image modeling pretraining on collective datasets. The SAM-AD is then adopted to extract robust features for the scene image, effectively reducing the impact of noise in road scenes. In unsupervised domain adaptation, it is suboptimal to alleviate the domain gap between the source and target domain by simple alignment. \cite{learntoadapt} propose to utilize SAM's feature space as a robust supervision, and both explicitly align source and target domain with SAM in 3D scene understanding task.

Distill LLM knowledge to road scenes. LLM pretrained large-scale datasets is regarded as a rich knowledge base. For example, DriveMLM\cite{drivemlm} is an LLM-based autonomous driving framework that inputs multi-modal data observed on the surrounding scene, and then employs a MLLM to model behavior planning. Behavior planning involves determining the optimal driving route based on the surrounding environment and giving decision states for vehicle control. Here, the teacher model is an existing AD system such as Apollo\cite{apollo}. The DriveMLM's linguistic decision outputs are aligned with Apollo's decision state output by its behavioral planning module, transforming them into formats that can be easily processed by MLLMs. 

\textbf{Instant learning }addresses the necessity for autonomous driving systems to instantly adapt to previously unseen scenarios. Inspired by the ability of humans to instantly learn new things through just a few examples, the models for road scene understanding should be able to adapt and learn from these new scenes or tasks with few shot exemplars. Moreover, when faced with a new situation, the model can instantly learn to generalize with rich instructions, ensuring the safety of intelligent vehicles in an ever-changing real-world environment. Two effective approaches to instant learning are in-context learning and instruction tuning. 

In-context learning, particularly referring to few-shot prompting, involves the rapid adaptation of the perception system to a new environment based on few-shot example pairs, analogously to GPT-3\cite{gpt-3}. This rapid adaption involves scenario adaption and context adaption, simultaneously. For instance, instead of extensive retraining to numerous instances of this scenario, the system can learn and adapt quickly based on a few specific examples when encountering a new scenario on the road. Furthermore, the driving system leverages few-shot examples to make rapid and contextually informed decisions in real time as it faces similar situations. Typically, Dolphins\cite{dolphins} is the pioneering work that harnesses the in-context capability of VLMs\cite{openflamingo} to instantly learn and adapt to a series of driving-related tasks. Extensive experiments on DriveLM\cite{drivelm} have demonstrated instant learning and adaptation capability in prediction, planning, and reasoning tasks through in-context learning.

Instruction tuning has also emerged as an effective technique to enhance generalization abilities of FMs like \cite{instructblip,instructgpt}. To unlock the powerful capabilities of FMs and apply them to solve driving tasks, researchers like \cite{dolphins} have attempted to transform road-related data in input-output tuples into instruction-following data organized in instruction-input-output triplets. These instructions may be task-specific descriptions or requirements annotated by an efficient auto-machine like ChatGPT. Combined with in-context learning, self-instruct tuning is also a simple technique to align the foundational capability of FMs with driving intention.

In conclusion, knowledge distillation facilitates the efficient transfer of foundational knowledge of FMs, while instant learning enables adaptability to new road scenes. Both are crucial components in the development of robust, adaptable, and efficient road scene understanding.

\subsection{Continual Learning}
In a real-world application scenario where sustainability is the key concern, we naturally expect the model to behave like humans, adapting to new tasks continually. This gives rise to the study of continual learning, where the pretrained model is required to continually learn a sequence of new tasks without forgetting previously learned ones. Contrary to traditional machine learning models which are predicated on the notion of encapsulating a static data distribution, continual learning is more vulnerable to \textit{catastrophic forgetting}\cite{catastrophic_forgetting,rehearsal1_icarl,regulairzation2_ewc,architecture3_pnn}, where the acquisition of new knowledge leads to an abrupt erosion of previously learned information. 


\textbf{Rehearsal-based methods} are often preferred when applying continual learning methods to task-specific road scene systems. To adapt the pretrained model to various weather conditions, Liang et al.\cite{brain_domain_incremental} introduce a novel rehearsal strategy in domain-incremental learning, employing a two-stage ``Recall" and ``Adapt" process. The ``Recall" stage serves as a rehearsal mechanism, utilizing self-training with mixed domain data to search and reinforce previously learned domain characteristics, while the ``Adapt" stage incrementally introduces new domains, utilizing patch-based adversarial learning to fine-tune the discriminability and generalizability of the model. DISC\cite{weather_DIL} proposes a model that can incrementally learn to detect things in different weather conditions. It stores the statistical parameters and simply "plug and play" the statistical vectors for the corresponding task into the model in new tasks, without the need for expensive feature banks. 

\textbf{Architecture-based methods}. Traditional continual learning methods usually adopt an expandable architecture to accommodate new tasks automatically. DRB\cite{DRB} presents a novel approach utilizing a neuro-fuzzy architecture called Deep Rule-Based (DRB) system to integrate the malleability of fuzzy logic to adapt to new data continuously. DRB evolves its architecture by comparing incoming data against existing prototypes using a distance metric, adding new rules, or updating existing ones based on predefined thresholds. Concurrently, it modifies its meta-parameters, such as rule shapes and decision thresholds, ensuring that the model continuously refines and adapts its decision-making process to the ever-changing data landscape.

\textbf{LLMs as knowledge base.} Benefiting from the extensive pretraining of LLMs, perception models based on LLMs generally demonstrate the ``Emergent ability"\cite{emergent}, which refers to the phenomenon where these models exhibit capabilities not explicitly programmed or anticipated during their training. This ability, along with the transfer ability discussed in \ref{section:transfer_ability}, enables the model to ``learn" new prototypes more quickly and accommodate new scenarios more efficiently\cite{gpt-4v}. Given this kind of ability, in the trend of merging LLMs into autonomous systems, existing systems\cite{drivesafe, 3R, dilu, drive_as_you_speak, drivelikehuman} mainly follow the ``store-and-search" manner. In this manner, the model uses an explicit natural language knowledge base to cache past skills and search for useful information when it encounters a similar situation. Fu et al.\cite{drivelikehuman} propose a knowledge base storing the expert advice that deviates from the original pretrained model's knowledge. This expert advice helps the model make more proper and practical decisions just like human drivers do. Next time it encounters the same situation, the model will be reminded of the taught expert knowledge by the in-context learning mechanism\cite{gpt-3}. DiLu\cite{dilu} designs a vector knowledge base taking vectorized scene descriptions as key and searches for related information by cosine similarity. It further designs a reflection module to identify whether the decision is safe, subsequently refining unsafe decisions into safe ones using the knowledge embedded in the LLM and updating these revised decisions into the knowledge base. 

We anticipate seeing works in the realm of LLMs or VLMs for road scenes that align with the classical paradigm of continual learning in the future. In fact, within the domain of general-purpose LLMs or VLMs, numerous parameter-efficient fine-tuning (PEFT) methodologies can, to a certain extent, alleviate the issue of catastrophic forgetting, such as LoRA\cite{lora} and various Adaptors\cite{adalora, llamaAdaptor}. We hope that future research will integrate some of the unique aspects of the road scene (corner case\cite{li2022coda} and long-tail problem\cite{long-tail}) and propose knowledge-enhanced continualable fine-tuning methods. By tailoring these advanced fine-tuning techniques to the distinctive challenges of the road scene, we can enhance the model's ability to adapt and retain crucial information, ensuring more robust and reliable performance in road-scene understanding applications.

\subsection{Learn to Interact}

Although autonomous systems in road scenes are capable of collecting various types of data from the environment, such as images and radar points, it is desirable for them to also consider other environmental factors, including the driver's emotions or instructions. Road scenes are inherently dynamic, with continuous changes in traffic patterns, pedestrian movements, and environmental conditions, necessitating a system that can capture and respond in real-time. The more elements the system takes into account from its environment and the more frequently it interacts with these elements, the closer it approaches the driving habits of humans and the safer it will be. In this subsection, we will discuss the interaction of autonomous driving systems with their environment, and how these interactions contribute to the development of robust and safe autonomous systems.

\textbf{Interaction with human.}
Nobody's going to leave their safety to a black box. There are many works \cite{multitask-8, multitask-7} generating additional reasoning explanations for each control/action decision which help users understand the current state of the vehicle and the surrounding environment. To enable continuous conversation, LLM-based models such as DriveGPT4\cite{drivegpt4},  LMDrive\cite{lmdrive}. build vast instruction datasets on road scenes and train the model in an instruction-tuning manner.

\textbf{Error recovery.}
Despite the robust reasoning capability of LLMs, LLM-based autonomous systems are still vulnerable to some intricate cases such as a really rare object in the picture, poor feature quality due to the perception sensors, or intrinsic hallucination problem\cite{hallucination}. These challenges are particularly pronounced in road scenes, where unexpected obstacles or conditions can arise at any moment. Dolphins\cite{dolphins} proposes a reflection mechanism to enable the model to correct the mistakes by itself through continuous conversations with users. LMDrive \cite{lmdrive} proposes a dataset where notices from humans are included. These occasional notices during driving, such as ``Watch out for the red traffic light in front of you" are appended to the corresponding frame tokens and fed into the LLMs for reasoning. These notices, along with navigation instructions from navigation apps, are designed to simulate real-life scenarios where a passenger or a side assistance system communicates with the driver to navigate safely through dynamic and potentially hazardous road environments. 

\textbf{Adaptive driving behavior.}
In road scenes, the driving behaviors should be taken into consideration for safe driving\cite{weiwei2024learning}. The complex interplay between a driver's emotional state and the unpredictable nature of road scenes can significantly influence driving behavior. For example, the ego car should slow down when detecting fright or horror from the driver's expression which may be caused by some potential risks that autonomous systems fail to detect. Sonia et al.\cite{emotion_1}introduce a methodology that employs in-vehicle sensors to identify mild cognitive impairment signals in elderly drivers in real-time. Upon detection, the system alerts the driver and offers driving assistance, thereby ensuring safe driving behavior and reducing the risk of accidents amidst the unpredictable dynamics of road scenes. Moreover, the driver's emotions will influence the trajectory prediction system. For example, an angry driver will lead the vehicle trajectory to show a more severe lateral deviation and a more violent longitudinal acceleration\cite{angery_driver}, which is difficult for traditional models to predict. CPSOR-GCN\cite{predict_cpsor} proposes a trajectory model that integrates both cognitive and physical features and uses stimulus-organism-response theory to model abnormal emotions. The experiment verifies that the incorporation of emotional data helps enhance prediction accuracy.

In conclusion, by enabling sophisticated interactions with human inputs, facilitating error correction through reflective mechanisms, and incorporating adaptive behavior based on emotional and situational awareness, these models contribute to the development of more intuitive and responsive autonomous vehicles. Further research into these areas is essential to address remaining challenges and to fully leverage the capabilities of LLMs or VLMs in improving the performance and trustworthiness of road-scene autonomous systems.

\subsection{Generative Foundation Models}
Generative models leverage sophisticated algorithms to model the complex patterns and relationships within visual data. They represent the full probability distribution of all variables, modeling the joint probability distribution of both current and target variables. Consequently, a generative model can simulate or generate the distribution of any variable. They employ various architectures such as generative adversarial networks (GANs), variational autoencoders (VAEs), autoregressive models, or diffusion models to achieve this objective.
 
In the context of road scene understanding, generative foundation models\cite{zheng2024genad,su2024text2street,world_model,recurrentworldmodel,UniWorld,gaia1,drivedreamer,trafficbots,muvo,jia2023adriveri,recore} refers to the ability of the model to imagine and generate a sequence of upcoming scenarios and actions when given the past scenarios and actions, empowering the driving system to adapt to changing environments and making safe actions.
Generative world models achieve this predictive capability through leveraging various techniques such as representation abstraction, sequential modeling, auto-regressive prediction, or more advanced architectures. These techniques can all capture temporal-spatial and long-term dependencies across video frames. Modeling the dynamic and structural world is a challenging task, especially easily overlooked details such as traffic lights, so recently many efforts have been made to build a generative world model.

\textbf{Image representation}. Sora\cite{sora} recently has attracted great attention due to its ability to follow human instructions to generate high-quality videos with highly spatial and temporal consistency. GAIA-1\cite{gaia1} as a generative world model, seamlessly integrates video, action, and text inputs, enabling fine-grained control and comprehensive understanding of complex environments. The world model processes discrete tokens from input as sequence modeling and then autoregressively predicts the next scenario tokens. It shows its controllable driving videos by passing the text prompt, such as a minimal change(turn the red light to green). Besides, introducing road structural information as an auxiliary condition is also an effective method for understanding road scenes. DriveDreamer\cite{drivedreamer} is a real-world-driven generative world model, excelling in controllable video generation and future driving action generation. To improve sampling efficiency, the diffusion-based world model utilizes structural information (such as HDMaps, 3D Boxes) as guidance to efficiently sample in an extensive search space. These methods are learned in 2D representation space such as RGB images and videos. Although RGB images can capture rich information such as context, color, and shape information, they might struggle with spatial structures. 

\textbf{Occupancy representation}. OccWorld\cite{predict_occ_multimodal_1} proposes a novel multi-task world model for joint future scenario generation and ego trajectory prediction in 3D occupancy representation. Unlike predicting a single token at each time like GPT-3, they propose a novel spatial-temporal transformer as the world model, which can simultaneously generate multiple tokens each time. A U-shape network is also adopted to generate multi-scale scenario tokens and aggregate them for next token prediction. Structural understanding and easy accessibility of occupancy representation allow this approach to accommodate large-scale training.

It is worth considering how to learn a world model keeping a connection with existing FMs, and finally promote realistic applications in real-world scenarios.  Existing generative world models that predict the coming states (including scenarios and actions) conditioned the past states have significant limitations of real-time interaction with the environment, leading to a relatively poor performance and inexplicable generation. Besides, simple action space which merely involves limited levels of steers and speed, is also far from meeting practical needs. Instead, constructing a close-loop generative world model is a promising trend and more in line with reality, where the model keeps real-time interaction with the environment. Furthermore, defining various action spaces\cite {voyager} and designing a novel search strategy \cite{voyager,diversity} are both useful ways to deal with real-world challenges.

\section{Datasets and Evaluation Metrics}
\label{sec:datasets}
\subsection{Datasets}
\begin{table*}[t]
    \centering
    \adjustbox{width=\textwidth}{
        \begin{tabular}{*{12}{c}}
        \toprule
         \multirow{2}{*}{\textbf{Datasets}} & \multirow{2}{*}{\textbf{Statistics}} & \multirow{2}{*}{\textbf{Scenario}} & \multicolumn{4}{c}{\textbf{Task}} & \multicolumn{5}{c}{\textbf{Modality}} \\  \cmidrule(lr){4-7} \cmidrule(lr){8-12} \\ 
         & &  & Perception& Prediction &Planning&Understanding& Image&Video&Text&Action&Point Cloud\\ \midrule
         LaMPilot\cite{lampilot} &4.9k scenes, 7.6 instruction length per frame & simulator&\XSolidBrush&\XSolidBrush&\checkmark&\XSolidBrush &\checkmark &\checkmark &\checkmark &\checkmark &\XSolidBrush    \\ \midrule
         DriveLM-nuScenes\cite{drivelm}&4,871 frames, 91.4 QAs per frame& urban&\checkmark&\checkmark&\checkmark&\checkmark &\checkmark &\XSolidBrush &\checkmark &\XSolidBrush &\XSolidBrush   \\ \midrule
         
         DriveLM-CARLA\cite{drivelm}&183,373 frames, 20.5 QAs per frame& simulator & \checkmark&\checkmark&\checkmark&\checkmark &\checkmark &\XSolidBrush &\checkmark &\XSolidBrush &\XSolidBrush   \\ \midrule

         DriveGPT4\cite{drivegpt4}&28K video, 16K fixed QAs, 12K conversations& urban,rural &\checkmark&\checkmark&\checkmark&\checkmark &\XSolidBrush &\checkmark &\checkmark &\checkmark &\XSolidBrush   \\ \midrule

         Talk2BEV\cite{talk2bev}&1k BEV scenarios, 20 QAs per scenarios & urban  &\checkmark&\XSolidBrush&\XSolidBrush&\checkmark &\checkmark &\XSolidBrush &\checkmark &\XSolidBrush &\checkmark   \\ \midrule
         
         Rank2Tell\cite{rank2tell}&116 scenarios, 31.95 caption length per scenarios& urban &\checkmark&\XSolidBrush&\XSolidBrush&\checkmark &\checkmark &\checkmark &\checkmark &\checkmark &\checkmark   \\ \midrule

         NuPrompt\cite{languageprompt}&34k frames, 1.1 prompts per frame& urban & \checkmark&\checkmark&\XSolidBrush&\XSolidBrush &\checkmark &\XSolidBrush &\checkmark &\XSolidBrush &\XSolidBrush   \\ \midrule
         NuScenes-QA\cite{nuscenes-qa}&30k frames, 15.3 QAs per frame& urban & \checkmark&\XSolidBrush&\XSolidBrush&\checkmark &\checkmark &\checkmark &\checkmark &\XSolidBrush &\checkmark   \\ \midrule

         DriveMLM\cite{drivemlm}&50k routes, 30 scenarios, 200 trigger points per scenarios& simulator&\checkmark&\checkmark&\checkmark&\checkmark &\checkmark &\XSolidBrush &\checkmark &\checkmark &\checkmark   \\ \midrule
            
         LMDrive\cite{lmdrive}&64K parsed clips, 464K notice instructions& simulator&\checkmark&\checkmark&\checkmark&\checkmark &\checkmark &\checkmark &\checkmark &\checkmark &\checkmark   \\ \midrule

         Reason2Drive\cite{reason2drive}&600K video-text pairs & urban,rural&\checkmark&\checkmark&\XSolidBrush&\checkmark &\checkmark &\XSolidBrush &\checkmark &\XSolidBrush &\XSolidBrush   \\ \midrule

         Driving-With-Llm\cite{drivingwithllm}&10k driving scenarios, 16 QAs per scenarios& urban&\checkmark&\checkmark&\XSolidBrush&\checkmark &\XSolidBrush &\XSolidBrush &\checkmark &\XSolidBrush &\XSolidBrush   \\ \midrule

         Refer-KITTI\cite{referring}&6.6k frames, 10.7 instances per prompt& urban,rural&\XSolidBrush&\XSolidBrush&\checkmark&\checkmark &\XSolidBrush &\checkmark &\checkmark &\checkmark &\XSolidBrush   \\ \midrule

         NuScenes-MQA\cite{NuScenes-MQA}&34k scenarios, 41.2 QAs per scenarios& urban&\checkmark&\XSolidBrush&\XSolidBrush&\XSolidBrush &\checkmark &\XSolidBrush &\checkmark &\XSolidBrush &\XSolidBrush   \\ \midrule

         LiDAR-text\cite{lidarllm}&420K 3D captioning data, 280K 3D grounding data& simulator & \checkmark&\XSolidBrush&\XSolidBrush&\checkmark &\XSolidBrush &\XSolidBrush &\checkmark &\XSolidBrush &\checkmark   \\ \midrule

         CARLA-NAV\cite{carla-nav}&83k frames, 7 command length per frames& simulator & \XSolidBrush&\XSolidBrush&\checkmark&\XSolidBrush &\checkmark &\XSolidBrush &\checkmark &\XSolidBrush &\XSolidBrush   \\ \midrule

         RSUD20K\cite{rsud20k}&20K high-resolution, 130K bounding box annotations& urban & \checkmark&\XSolidBrush&\XSolidBrush&\XSolidBrush &\checkmark &\XSolidBrush &\checkmark &\XSolidBrush &\XSolidBrush   \\ \midrule

         DRAMA\cite{multitask-7}&18k scenarios, 5.6 text strings per scenarios& urban & \XSolidBrush&\XSolidBrush&\checkmark&\checkmark &\XSolidBrush &\checkmark &\checkmark &\checkmark &\XSolidBrush   \\ \midrule
        \end{tabular}
    }
    \caption{Summary of up-to-date language-based multi-modal multi-task datasets for road scene understanding. The summarized perspectives of these datasets involve considerations such as data statistics, applicable scenarios, tasks, and annotated modalities.}
    \label{table:datasets}
    
\end{table*}
The exploration of language-guided visual understanding for road scenes indeed stands at the forefront of advancements in intelligent vehicles. Pioneering studies in this domain go beyond single-task applications, aiming to create more versatile and adaptable systems through the integration of language guidance across multiple tasks and modalities. As presented in Table.~\ref{table:datasets}, there is a summary of up-to-date language-based multi-modal multi-task datasets specifically designed for road scene understanding. It's noteworthy that these datasets depart from traditional ones, like CODA\cite{li2022coda}, KITTI\cite{kitti}, and Cityscapes\cite{cityscapes} which typically annotate road scenes solely with geometric labels, such as instance-level bounding boxes and pixel-level masks. The absence of semantic and contextual information in traditional datasets poses a great challenge to achieving intelligent driving systems.

The shift towards language-guided multi-modal multi-task datasets implies a richer and more context-aware annotation approach. Instead of relying solely on geometric annotations, these datasets incorporate language-based descriptions or instructions that guide the understanding and reasoning of the scene image, encompassing multiple tasks and modalities to forge more versatile and adaptable models. 
Annotations in these datasets are provided in the form of question-answer pairs (QAs),  either collected from new scenes like DRAMA\cite{multitask-7} or extended on existing datasets like DriveLM\cite{drivelm}.  
Moreover, there is a trend that QA pairs for different tasks are organized from Chain-of-Thought like Reason2Drive\cite{reason2drive}, Tree-of-Thought like NuScenes-QA\cite{nuscenes-qa} and Graph-of-Thought like DriveLM\cite{drivelm}.

\subsection{Evaluation Metrics}

Existing MM-VUFMs are applicable to a wide range of tasks, so it is necessary to summarize some representative evaluation metrics. Here, we briefly introduce commonly used evaluation metrics designed to measure the quality and diversity of the data generated by MM-VUFMs from the perspective of modality generation.

For text generation tasks, such as visual question answering (VQA) and image captioning, metrics like BLEU and METEOR are widely adopted, serving to evaluate the quality of the generated texts by comparing them against reference texts.
For tasks involving image generation, the Fréchet Inception Distance (FID) is a metric used to assess the quality and diversity of generated images. It quantifies the quality of generated images by examining the distance between the feature distributions of generated images and real images. Similarly, the Frechet Video Distance (FVD) is employed to evaluate video generation tasks, analogous to FID for images. FVD evaluates the quality and diversity of generated videos by measuring the discrepancy in feature distributions between the synthetic videos and real videos. For action generation tasks, such as simulating driving actions and steering angle, typical metrics are adopted like L1 and L2 errors.

Besides, for specific vision tasks in the context of perception, prediction, we use specialized metrics such as mean average precision (mAP) for object detection, mean intersection over union (mIoU) for semantic segmentation, average end point error (ADE) for trajectory prediction, and etc.

These metrics can be used individually or jointly to comprehensively evaluate diverse MM-VUFMs.

\section{Open Challenges \& Future Trends}
Within the promising landscape of road scene understanding, numerous challenges persist. Identifying and addressing these challenges are crucial to ensure efficient and reliable autonomous driving systems. In this section, we delve into open challenges shared by existing approaches and also highlight future trends that reveal insights toward solving these challenges.

\begin{figure}[t]
    \centering
    \includegraphics[width=0.8\linewidth]{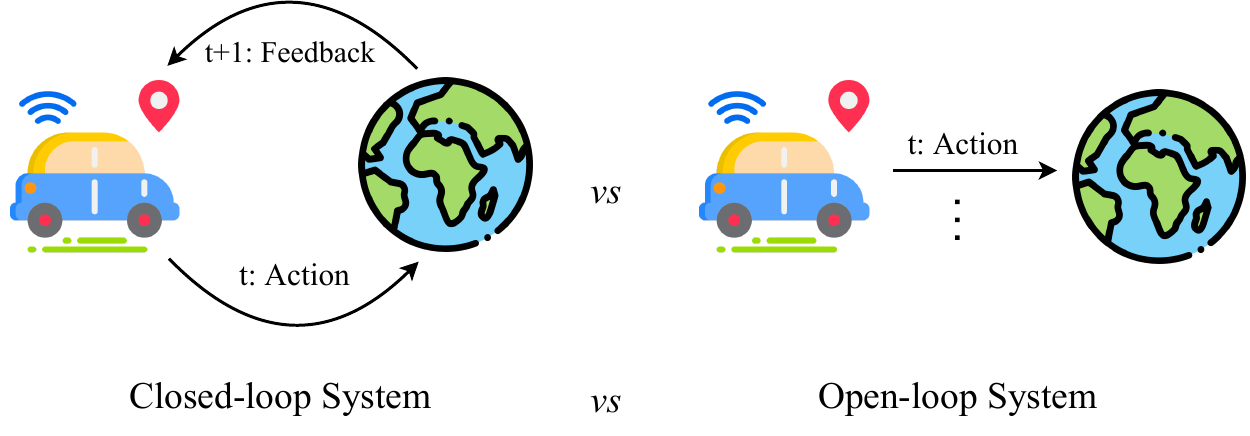}
    \caption{Closed-loop system vs Open-loop system. In the closed-loop system, it maintains real-time interaction with the surrounding environment. This involves executing actions to modify the environment at the current time \(t\) and receiving feedback from the environment at the next time \(t+1\). In the open-loop system, actions are undertaken at each time, but no feedback is received from the environment.}
    \label{fig:closed-loop}
\end{figure}
\subsection{Closed-loop vs Open-loop Driving System}
The concepts of closed-loop system and open-loop system play a crucial role in control systems. 
An open-loop autonomous driving system operates without real-time feedback from the environment. They execute predefined programs or instructions without paying attention to external changes. In contrast, their behavior or reaction in the closed-loop system depends on the real-time feedback observed from sensors and the environment. Compared to the open-loop system, the closed-loop system is generally considered more reliable and safer for real-world road scenes, as illustrated in Fig.~\ref{fig:closed-loop}.

Among visual understanding foundation models for road scenes, there is a promising trend from single-modal to multi-modal input. Coupled with the language modeling ability of LLM, existing approaches mostly use LLM as a unified multi-modal interface by tokenizing multi-modal input into language tokens and making next token prediction for understanding and reasoning. However, these approaches operate in an open-loop setting. More specifically, these approaches are usually trained and evaluated on assumed benchmarks such as commonly used nuScenes\cite{nuScenes} with already predefined properties. This pipeline brings two primary challenges: 1) restricted exposure to real-world data and  2) failure to incorporate feedback signals from a dynamic environment. Recently, researchers have started to explore closed-loop driving systems. LMDrive\cite{lmdrive} is the first closed-loop driving framework that takes language instruction and multi-sensor data as input and output control actions, based on real-time observation from real-world road scenes, but there is still limited performance to achieve fully closed-loop driving systems.

To bridge the gap between unrealistic open-loop and the reliable closed-loop driving agent,
driving simulator is a promising research trend. It serves as a customized and flexible platform to expose proposed approaches to 3D driving scenes and mimic closed-loop evaluation with real-time feedback. CARLA\cite{carla}, known as a commonly used open-source simulator for autonomous driving, has become a cornerstone in the development and evaluation of autonomous driving systems. But existing simulators\cite{unisim,carla} still have problems of lacking fidelity and limited real-world reappearance. These simulated traffic participants like pedestrians have significant differences in appearance compared with real traffic participants, potentially impacting the performance of algorithms trained solely on simulated data. Besides, real sensors face various environmental factors (like various weather conditions and occlusions) that might not be fully captured in simulations and the simulators struggle to reproduce large-scale environments (like complex intersections) with the same level of detail and accuracy. Therefore, developing a simulator that can simulate high-quality driving scenes is a promising direction.

\subsection{Interpretability}

Interpretability in road scene understanding is indispensable, as it ensures safety and accountability, and enables effective debugging along with continuous improvement. Interpretable foundation models show not only what they respond, but also the reasons behind these responses.

In this direction, existing algorithms\cite{multitask-8,multitask-7,multitask-6} have investigated the use of FMs (LLMs, VLMs) to explain their output in natural language. However, we contend that directly incorporating FMs, which are used as the text generation head, into the algorithm framework cannot ensure interpretability. Although excelling at language tasks, they operate as black boxes due to their internal reasoning process remains unclear. Therefore, Chain-of-Thought (CoT) has been proposed to visualize the multi-step reasoning of LLMs to simulate the human thinking process. By depicting how LLMs process inputs and visualize outputs, this technique tries to unmask internal thought for enhancing interpretability. 

It is also greatly crucial to make reasonable actions for end-to-end autonomous driving systems. Although they often employ simpler architecture than modular design and adopt global optimization objectives, it is difficult to explain. Some post-hoc techniques can be adapted to enhance the interpretability of driving models such as saliency maps\cite{multitask-6,multitaskwithAttention}, decoding intermediate outputs into results. The output of interpretable tasks such as object detection\cite{transfuser,multimodalFusiontransformer}, depth estimation\cite{hidden,transfuser} is another effective way to enhance interpretability.   
Recently, there is also an emerging solution that transforms driving tasks into visual question answering to explain the decision-making process\cite{drivelm,explaining}.

\subsection{Low-resource Condition}
Low-resource condition usually exists in road scenes where there is a great challenge of limited high-quality data, insufficient computing resources and memory capacity on an embedded intelligent vehicle. Early methods have attempted to deal with this challenge via model quantization\cite{xu2019training,quantized}, memory-efficiency cache technique\cite{bechtel2018deeppicar}, robust post-processing step\cite{sabater2020robust} and edge-cloud cooperation\cite{edge}, respectively. They significantly reduce latency time and save memory capacity under the constraint of low-resource condition, ensuring safe and efficient driving.


In the new era of foundation models, although their generalization capability allows easy adaptation to various domains and tasks, they struggle to generalize well under some low-resource conditions due to limited data, fine-grained differences, and highly specialized domain, as reported in \cite{zhang2024low}. These characteristics also exist in road scenes where large-scale high-quality data are difficult to collect, and there is a very subtle appearance difference between traffic objects. Obviously, simply reducing model parameters or improving model efficiency like above methods\cite{xu2019training,quantized,bechtel2018deeppicar,sabater2020robust,edge} fail to work well. Many researchers have investigated many learning methods to overcome these new challenges, e.g., zero-shot learning\cite{zhang2023resimad,yang2024diffusion}, few-shot learning\cite{mei2023few,liu2023generalized}. They aim to achieve good performance with only few training exemplars or even without any training exemplars. To differentiate fine-grained details across objects with different semantics and similar appearances, support examples or templates in few-shot learning\cite{wang2020frustratingly} seamlessly benefit distinguishing them. FOMO\cite{zohar2023open} provides a novel perspective to use natural language predicted by LLM to describe fine-grained differences to address the challenge of open-world detection in real-world scenarios. Generative methods are also effective ways to alleviate data scarcity via generating diverse and unlimited data, which make controllable and human-friendly condition generation a reality. These methods are mainly based on diffusion models\cite{yang2024diffusion,ling2023align}, road scene simulation\cite{wei2024editable,wu2023mars,zhang2023nerf,chen2024s}, generative gaussian splatting\cite{zhou2023drivinggaussian}, respectively. Moreover, some parameter-efficiency techniques \cite{wang2024revisiting,lee2024bayesian,hayati2024chainofinstructions} are proposed to adapt foundation model to downstream scenarios with only finetuning a few trainable parameters. Although great progress have been made, performance gain is very limited, still existing a long way to go before foundation models achieve good generalization under the low-resource condition.

\subsection{Embodied Driving Agent}
Although visual understanding FMs can easily incorporate sensor data, model the realistic world, and complete driving-related tasks according to human instruction, these models are pretrained on given the input data collected and produced by humans without actively solving unknown tasks and polishing their own behaviors. In contrast, we argue that transforming visual understanding FMs into embodied driving agents is a promising trend to achieve DriveAGI by endowing them with embodied reasoning capacity.

Recent advancements in LLM and multi-modal learning have brought the concept of embodied intelligence into practical application, such as robotics. PaLM-E\cite{palme}, RT series\cite{RT-1,rt-2,RT-X} have begun to explore generic robotic agents with embodied reasoning capabilities such as mobile manipulation and task/motion planning.
However, having similar embodied reasoning capabilities in driving systems encountered obstacles. Unlike splitting a large task into smaller ones and then progressively tackling them as they typically do in robotic tasks, driving outdoors is a highly coupled and complex task that cannot be step-by-step solved by simply task decomposition strategy. Moreover, the uncertainty associated with driving behavior poses a significant challenge. In this direction, a feasible solution is to be in line with end-to-end driving framework based on reinforcement learning. To imitate human driving behavior, the system collects multi-sensor data, maps them into low-level actions with the global optimization objective, and calibrates or stimulates their next actions with the feedback path. Surprisingly, ELM\cite{zhou2024elm} has recently taken a solid step towards achieving embodied driving agent.

\subsection{World Model}
The world model\cite{wang2023driving,driveworld} refers to the model that makes reasonable predictions given past scenarios and actions, which can not only generate controlled, diverse, and scalable samples for training robust driving systems, but also promote them to perfectly generalize real-world situations.

Modeling complex driving scenarios is a great challenge due to small yet important details such as traffic lights, which are easily overlooked. Existing methods such as GAIA-1\cite{gaia1} and DriveDreamer\cite{drivedreamer} harness the remarkable modeling capability of the diffusion model but still face the challenge of sample inefficiency, leading to slow convergence and a great requirement for computing resources. 
Considering that they only model sequences with lengths of thousands, LWM\cite{liu2024world} proposes the RingAttention technique to extend the sequence length to millions, enabling broader capability in understanding the world. In addition, we contend that the simple action space (such as speed and steering) cannot be sufficient to meet the real-world driving requirement since human drivers can make fine-grained controls when facing unpredictable tasks. In embodied reinforcement learning, Voyager\cite{voyager} is designed to progressively solve open-ended tasks in Minecraft Game using a high-dimensional action space in code rather than low-level motion commands. These action programs are not generated at one time but require multiple iterations to mend until being verified for successfully completing the task. Similar inspiration could be implemented in the driving world model.

\section{Conclusion}
In this survey, we provide an overview of multi-modal multi-task visual understanding foundation models (MM-VUFMs) for road scenes. We systematically review the extensive literature focusing on task-specific models, unified multi-modal models, unified multi-task models and foundation model prompting techniques. Besides, we highlight their advanced strengths in diverse learning paradigms, involving open-world understanding, efficient transfer for road scenes, continual learning, interactive and generative capability. Finally, we also point out key challenges and future trends to push the boundaries of foundation models for road scene understanding.
In the new era of foundation models, we hope that this survey can provide researchers with comprehensive awareness and inspire future research in this domain.

\section*{Acknowledgement}
This work was supported by DiDi GAIA Research Cooperation Initiative (Grant No. CCF-DiDi GAIA 202304).

\bibliographystyle{plain}
\bibliography{ref.bib}
\textbf{Sheng Luo} is a Master student at Southeast University, Nanjing, China. His research interests include road scene understanding and computer vision.

\noindent\textbf{Wei Chen} is a Master student at Southeast University, Nanjing, China. His research interests include continual learning and computer vision.

\noindent\textbf{Wanxin Tian} received the Master degree in Electronics and Information Engineering from Beijing University of Posts and Telecommunications, Beijing, in 2021. He is currently a Senior Algorithm Engineer in DiDi. His research interests include computer vision, large language models and World Model.

\noindent\textbf{Rui Liu} is a Master student at Harbin Institute of Technology. Her research interests include visual generation and computer vision.

\noindent\textbf{Luanxuan Hou} is currently a senior algorithm engineer at Voager, DiDi Chuxing. His research interests include computer vision and diffusion model. He has published papers in CVPR, ICPR, and ICME and served as a reviewer for NeurIPS, ICML, and ICLR.

\noindent\textbf{Xiubao Zhang} is currently a senior expert algorithm engineer at DiDi Company. His research interests include computer vision, pattern recognition, and machine learning.

\noindent\textbf{Haifeng Shen} received the Ph.D. degree in signal and information processing from Beijing University of Posts and Telecommunications, Beijing, in 2006. He is currently a Principal Algorithm Engineer in DiDi. His research interests include computer vision, speech recognition, time sequence prediction and large language modeling.

\noindent\textbf{Ruiqi Wu} is a Master student at Southeast University, Nanjing, China. Her research interests include multi-task learning and computer vision.

\noindent\textbf{Shuyi Geng} is a Master student at Southeast University, Nanjing, China. Her research interests include continual learning and computer vision.

\noindent\textbf{Yi Zhou} is currently an associate professor with the school of computer science and engineering, Southeast University (SEU), China. His research interests include computer vision, pattern recognition, and machine learning. 

\noindent\textbf{Ling Shao} is the Founding CEO and Chief Scientist of the Inception Institute of Artificial Intelligence, Abu Dhabi, UAE. His research interests include Computer Vision, Deep Learning/Machine Learning, and Image/Video Processing. He is an Associate Editor of the IEEE TIP, the IEEE TNNLS, and several other journals. He is a Fellow of the IEEE, a Fellow of the IAPR.

\noindent\textbf{Yi Yang} is the Director of Technology Innovation Group at DiDi. His research interests include artificial intelligence, infrastructure and cloud native.

\noindent\textbf{Bojun Gao} is a Research Project manager at DiDi. His research interests include artificial intelligence and Educational Technology.

\noindent\textbf{Qun Li} is the Head of DiDi Research Outreach, is leading DiDi's efforts to work together with academic institutions worldwide, including research collaboration, talent cultivation and academic exchange. She was in the organizing committee of several workshops and tutorials. Her research interests lie in the intersection of artificial intelligence, autonomous driving and computer vision.

\noindent\textbf{Guobin Wu} is the Director of Technology Ecology and Development Department at DiDi. His research interests include artificial intelligence and computer vision.

\end{document}